\documentclass[journal]{IEEEtran}
\usepackage{silence}
\ifCLASSINFOpdf
\else
\fi
\usepackage[breaklinks, hidelinks]{hyperref}
\usepackage[justification=centering]{caption}

\usepackage{array}
\usepackage{amsmath,amsfonts,pifont}
\usepackage{algorithm}
\usepackage{algpseudocode}
\usepackage[subrefformat=parens,labelformat=parens,caption=false]{subfig}
\usepackage{textcomp}
\usepackage{stfloats}
\usepackage[breaklinks]{hyperref}
\usepackage{verbatim}
\usepackage{graphicx}
\usepackage{cite}
\usepackage[nolist,nohyperlinks]{acronym}
\usepackage{comment}
\usepackage{orcidlink}
\usepackage{adjustbox}
\usepackage{svg}
\usepackage{mathtools}
\usepackage[shortlabels]{enumitem}
\usepackage{soul}
\usepackage{colortbl}
\usepackage[nolist]{acronym}
\usepackage{bm}
\usepackage{balance}
\usepackage{soul}
\usepackage{adjustbox}
\usepackage{graphicx}
\usepackage{amssymb}
\definecolor{blu_2}{rgb}{0.0, 0.0, 0.9}
\usepackage[breaklinks, hidelinks]{hyperref}
\usepackage[justification=centering]{caption}
\usepackage{booktabs,tabularx,arydshln}
\usepackage{multirow,multicol}

\newcommand{\mycomment}[1]{}

\newcommand{\cmark}{\ding{51}}%
\newcommand{\xmark}{\ding{55}}%

\begin{acronym}
\acro{BiLSTM}{Bidirectional Long Short-Term Memory}
\acro{BCE}{Binary Cross Entropy}
\acro{CBAM}{Convolutional Block Attention Module}
\acro{CNN}{Convolutional Neural Network}
\acro{Conv-LSTM}{Convolutional Long Short-Term Memory}
\acro{DHP}{deep reinforcement learning-based head movement prediction}
\acro{DL}{Deep Learning}
\acro{DoF}{Degrees of Freedom}
\acro{DRL}{deep reinforcement learning}
\acro{FoV}{Field of View}
\acro{FSM}{Fused Saliency Map}
\acro{GBVS}{Graph-Based Visual Saliency} 
\acro{GCN}{Graph Convolutional Network}
\acro{GEP}{Gaze-based Field of View Prediction}
\acro{GNN}{Graph Neural Network}
\acro{GRU}{Gated Recurrent Unit}
\acro{HMD}{Head-Mounted Display}
\acro{KNN}{K-Nearest Neighbors}
\acro{LR}{Linear Regression}
\acro{LSTM}{Long Short-Term Memory}
\acro{ML}{Machine Learning} 
\acro{PoI}{Point of Interest}
\acro{QoE}{Quality of Experience}
\acro{RL}{Reinforcement Learning}
\acro{RMSE}{Root Mean Square Error}
\acro{RNN}{Recurrent Neural Network}
\acro{RoI}{Regions of Interest}
\acro{RR}{Ridge Regression}
\acro{SE-net}{squeeze-and-excitation network}
\acro{SE-Unet}{squeeze-and-excitation network and U-Net}
\acro{SP-ConvGRU}{Spherical Convolutional Gated Recursive Unit}
\acro{SP-GRU}{Spherical Gated Recurrent Unit}
\acro{SMSE}{Spherical Mean Squared Error}
\acro{S-SPCNN}{Spatial Feature Extraction module}
\acro{ST-SPCNN}{Saliency Feature Extraction module}
\acro{T-SPCNN}{Temporal Feature Extraction module}
\acro{VR}{Virtual Reality}
\acro{WLR}{Weighted Linear Regression}
\acro{CNN}{Convolutional Neural Network}
\acro{KL}{Kullback-Leibler Divergence}
\acro{CC}{Pearson Correlation Coefficient}
\acro{NSS}{Normalized Scanpath Saliency}
\acro{AUC-Judd}{Area Under the Curve–Judd}
\acro{SIM}{Similarity}
\acro{GAN}{Generative Adversarial Network}
\acro{CB}{Center Bias}

\end{acronym}

\begin{document}
\title{SalFormer360: a transformer-based saliency estimation model for 360-degree videos}

\author{Mahmoud Z. A. Wahba,~\IEEEmembership{Graduate Student Member,~IEEE,}
        Francesco~Barbato,~\IEEEmembership{Member,~IEEE,} Sara~Baldoni,~\IEEEmembership{Member,~IEEE,}
and~Federica~Battisti,~\IEEEmembership{Senior Member,~IEEE}% <-this % stops a space
\thanks{This work was partially supported by the European Union under the Italian National Recovery and Resilience Plan (NRRP) Mission 4, Component 2, Investment 1.3, CUP C93C22005250001, partnership on “Telecommunications of the Future” (PE00000001 - program “RESTART”)” and by the European Union’s
Horizon Europe Program under Agreement 101135637 (HEAT Project).}% 
\thanks{M. Z. A. Wahba, F. Barbato, S. Baldoni, and F. Battisti are with the Department of Information Engineering, University of Padova, Via Gradenigo 6b, 35131, Padua, Italy. (Corresponding author e-mail: \href{mailto:sara.baldoni@unipd.it}{sara.baldoni@unipd.it}).}
%<-this % stops a space
}


\markboth{Journal of \LaTeX\ Class Files,~Vol.~14, No.~8, August~2015}%
{Wahba \MakeLowercase{\textit{et al.}}: Bare Demo of IEEEtran.cls for IEEE Journals}


\maketitle

\begin{abstract}
Saliency estimation has received growing attention in recent years due to its importance in a wide range of applications. 
In the context of 360-degree video, it has been particularly valuable for tasks such as viewport prediction and immersive content optimization. 
In this paper, we propose SalFormer360, a novel saliency estimation model for 360-degree videos built on a transformer-based architecture. 
Our approach is based on the combination of an existing encoder architecture, SegFormer, and a custom decoder. 
The SegFormer model was originally developed for 2D segmentation tasks, and it has been fine-tuned to adapt it to 360-degree content. 
To further enhance prediction accuracy in our model, we incorporated a viewing center bias to reflect user attention in 360-degree environments. 
Extensive experiments on the three largest benchmark datasets for saliency estimation demonstrate that SalFormer360 outperforms existing state-of-the-art methods. 
In terms of Pearson correlation coefficient, our model achieves 8.4\% higher performance on Sport360, 2.5\% on PVS-HM, and 18.6\% on VR-EyeTracking compared to previous state-of-the-art.
\end{abstract}


% \fi
\IEEEpeerreviewmaketitle

\section{Introduction}

In recent years, \ac{VR} has gained widespread popularity, providing users with highly immersive experiences and allowing them to feel as if they were in a virtual world distinct from reality.

%Unlike traditional 2D screens, \ac{VR} allows users to freely explore the virtual environment they are immersed into. Users can freely move their head in the three directions, roll, pitch, and yaw (3DoF, three Degrees of Freedom), as well as move inside the virtual environment (6DoF, six Degrees of Freedom), effectively covering all possible motion. Furthermore, VR can enhance the immersive experience by enabling users to interact with the environment through hand tracking or any other techniques. 

Omnidirectional images and videos have become one of the most popular content types for \ac{VR}, thanks to the availability of user-friendly and low-cost 360-degree cameras. This content allows users to be placed at the center of a sphere and freely explore the environment in any direction by simply moving their heads. Although 360-degree media allows increased user immersion, their processing and transmission still entail many challenges. One major issue is the high memory and bandwidth requirements with respect to standard 2D content. For example, streaming a 4K 2D video requires approximately 25 Mb/s, while delivering a 4K resolution for each eye to provide a full 360-degree viewing experience requires around 400 Mb/s~\cite{ITU_743-10}. %Streaming platforms such as YouTube and Facebook typically send the entire 360-degree frame to the user. 

One effective way to address the transmission challenges of 360-degree videos is to implement a user-centered streaming paradigm. To this aim, human attention mechanisms have been studied  to design saliency estimation methods. These algorithms compute 2D probability maps which highlight the regions inside a 360-degree scene %that are 
most likely to draw users attention~\cite{BATTISTI2018}. These maps can then be used to transmit the salient regions at higher quality while encoding at lower quality (or discarding) less relevant areas~\cite{Alice2024}. 

Another possibility to reduce the transmission burden of 360-degree content considers the limitations of the human visual system, reflected by the \acp{HMD}. Indeed, despite the availability of 360-degree content, users only see a portion of the frame at a time, which constitutes about 20\% of the entire content~\cite{Nguyen2023}; the visible portion of a scene is commonly referred to as viewport. Viewport prediction can be integrated into streaming pipelines by delivering the predicted viewport in high quality, while downsampling or omitting peripheral regions. Based on this, different approaches have been proposed for predicting users' future viewport or fixation points. Specifically, past head and eye movements have been often employed as input and the integration of saliency information has shown to significantly contribute to the prediction~\cite{Rondón2022,Fan2020,Peng2023,Wang2022,Wang2024,Zhang2025,Setayesh_2024}. Indeed, compared to using head orientation alone, the integration of saliency maps enhances the accuracy of viewport prediction, enabling reliable forecasts over longer horizons (\textit{e.g.}, up to 5 seconds) as shown in~\cite{Rondón2022}.
These observations highlight that the proposed approach can be integrated in the transmission workflow for fostering efficient and effective streaming of 360-degree videos.

%Based on this, saliency estimation and other cues, such as user previous head orientations and, in some cases,  motion information, can be combined to predict the future viewports spanned by the users\cite{Rondón2022,Fan2020,Peng2023,Wang2022,Wang2024,Zhang2025,Setayesh_2024}. Viewport prediction can be integrated into streaming pipelines by delivering the predicted viewport in high quality, while downsampling or omitting peripheral regions. Compared to using head orientation alone, the integration of saliency maps enhances the accuracy of viewport prediction, enabling reliable forecasts over longer horizons (\textit{e.g.}, up to 5 seconds) as shown in~\cite{Rondón2022}.
%These observations highlight that accurate saliency estimation can foster efficient and effective streaming of 360-degree videos.

\begin{figure}[t]
    \centering
    %\graphicspath{{images/}}
    \includegraphics[width=.9\linewidth]{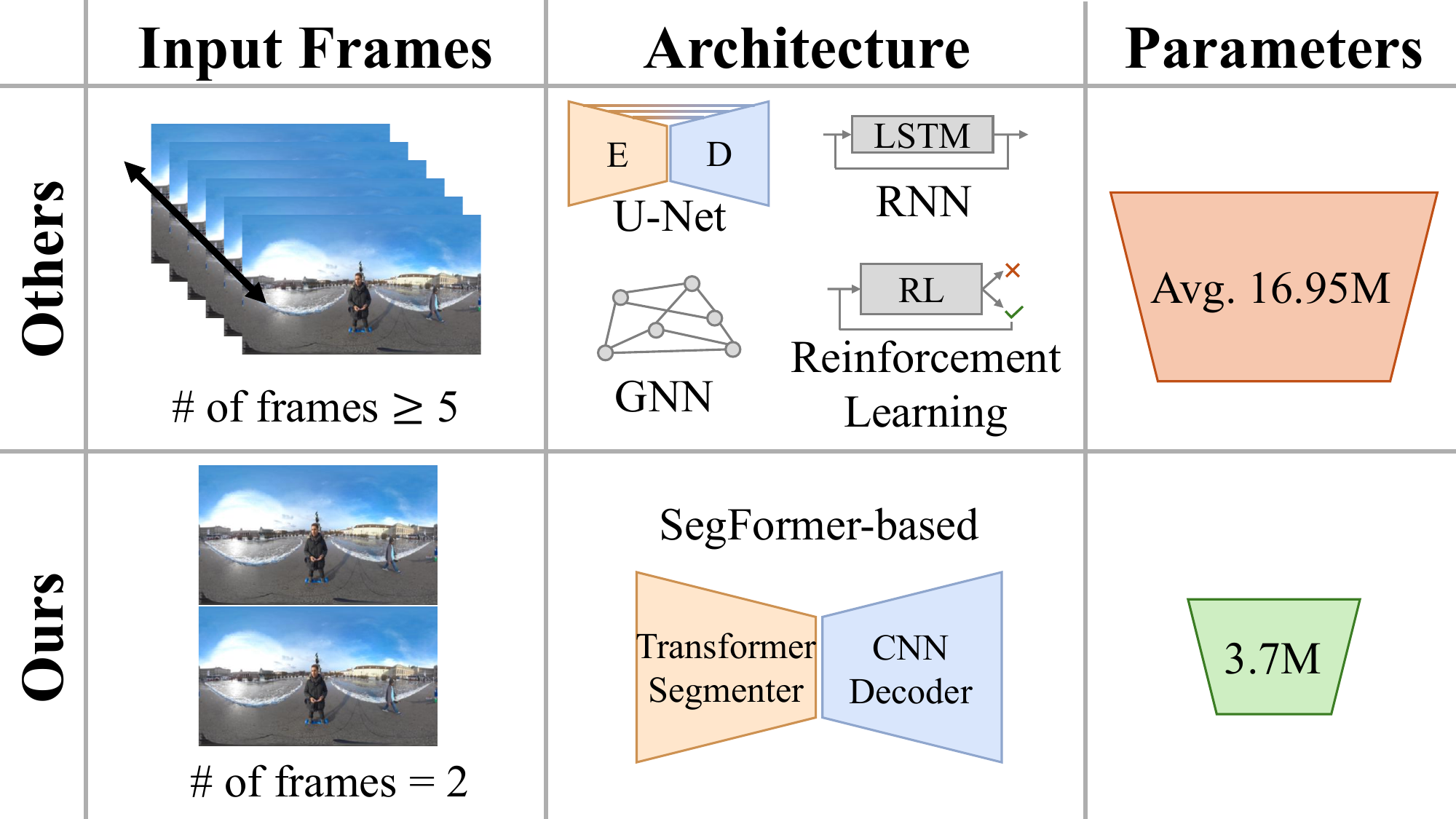}
    \caption{SalFormer360 can estimate future salient points in 360-degree videos using only a single previous frame and with limited computational resources. %Our approach improves the Pearson Correlation Coefficient by an average of $9.8\%$ across three datasets requiring $4.5\times$ less parameters.
    } 
    %\label{Segmentation results obtained by feeding 360-degree equirectangular frames into the SegFormer model (MiT-B0)}
    \label{fig:graphabs}
\end{figure}

% \begin{figure}[t]
%     \centering
%     %\graphicspath{{images/}}
%     \includegraphics[width=9cm]{images/abstract.pdf}
%     \caption{abstract graph} 
%     %\label{Segmentation results obtained by feeding 360-degree equirectangular frames into the SegFormer model (MiT-B0)}
%     \label{fig:abstract_graph}
% \end{figure}

Saliency estimation has been extensively studied in the context of 2D images and videos, where it plays a crucial role in applications such as object detection and video compression. With the rapid diffusion of immersive media, research in this area has now expanded towards 360-degree content. This shift presents new challenges due to the spherical geometry, larger field of view, and diverse viewing behaviors of users.

%Streaming 360-degree videos consumes a significant amount of bandwidth, which can lead to playback interruptions and negatively affect the user's \ac{QoE}. One effective solution to this issue is the use of saliency estimation models. By allowing streaming systems to allocate higher bitrates to the most salient regions and lower bitrates or even eliminate transmission for non-salient areas, we can optimize bandwidth usage.

Our approach (summarized in Figure~\ref{fig:graphabs}) is motivated by the observation that segmentation models are trained to detect and isolate objects within a scene, and saliency models tend to highlight regions of interest that often correlate strongly with those objects. We leverage this relationship by adopting the transformer-based, %pretrained 
2D segmentation encoder Segformer~\cite{Enze_2021} for our saliency estimation task. We feed 360-degree video frames into the Segformer encoder and exploit a custom decoder and viewing bias to generate the saliency maps.
%in processing, storage, and transmission. 

%In terms of processing, traditional techniques used for 2D image and video processing cannot be directly applied to 360-degree content. When 360-degree content is mapped to a 2D format using projection techniques such as Equi-Rectangular Projection (ERP) or Cube Map Projection (CMP), it experiences geometric distortions, particularly near the poles. 
%These distortions imply that particular attention must be paid when applying 2D processing techniques to 360 content, and ad-hoc approaches must be employed.  

%Regarding storage and transmission, 360-degree content requires more memory and higher bandwidth compared to standard 2D content. 

%Many methods have been proposed to predict this viewport and only transmit that specific section to the user, thereby enhancing the perceived quality of experience (QoE) while using bandwidth more efficiently. Machine learning and deep learning techniques have also been suggested for viewport prediction, with saliency maps being a key input generated through saliency estimation techniques.

%In this paper, we tackle the challenge of saliency estimation for 360-degree videos. 
Our contributions can be summarized as follows:

\begin{itemize}
   \item We introduce SalFormer360, a novel transformer-based model for saliency estimation. %This model is built on the 
   Building on the SegFormer backbone originally developed for 2D segmentation tasks, we fine-tune the encoder and design a custom decoder before training the whole model on 360-degree video datasets to produce high-quality saliency maps.

   \item We account for human exploration behavior by introducing a viewing bias to enhance the accuracy of saliency estimation.
   %introduce a viewing bias in our model to %better 
   %reflect user attention in 360-degree environments and to enhance the accuracy of saliency estimation.

   \item We %have prepared 
   extend %versions of 
   two 360-degree datasets: PVS-HM~\cite{Xu2018_2} and VR-EyeTracking~\cite{Xu2018}. We  generated ground-truth saliency maps from head orientation data and extracted RGB frames from MP4 videos, thus making them more accessible for future research in saliency estimation\footnote{Code and resources available at: https://github.com/LTTM/SalFormer360}.

   \item We conduct extensive experiments on the three largest benchmark datasets for 360-degree videos: Sport360~\cite{Zhang_ECCV_2018}, PVS-HM~\cite{Xu2018_2}, and VR-EyeTracking~\cite{Xu2018}. Our model outperforms state-of-the-art competitors, achieving an improvement in \ac{CC} of 8.34\% in Sport360, of 2.60\% in PVS-HM, and of 18.60\% in VR-EyeTracking compared to the best previously reported results.
\end{itemize}

The remainder of the paper is organized as follows: Section~\ref{sec:related_works} reviews related work %and the latest methods for
on saliency estimation. Section~\ref{sec:salformer} provides a detailed description of our proposed model. Sections~\ref{sec:results} and~\ref{sec:abltation_study} outline the experimental setup, the results, and provide ablation studies. Finally, Sections~\ref{sec:future_directions} and~\ref{sec:conclusions} discuss future directions and offer conclusive remarks.

\section{Related work}\label{sec:related_works}
%A saliency map is a visual representation which highlights the areas in an image or video frame that are likely to capture the user's attention~\cite{BATTISTI2018}. In recent years, there has been significant research in this area, that focuses on both traditional content and 360-degree content.
%Saliency estimation in 360-degree videos is a crucial task that aims to identify regions that are likely to capture the viewer's attention in a given scene~\cite{Baldoni_ISPA_2023}. This has numerous applications, including video compression and viewport prediction. In the context of viewport prediction, saliency estimation can serve as an essential input to enhance accuracy and extend prediction horizons. By incorporating predicted saliency maps alongside other inputs - such as previous head positions - viewport prediction frameworks can achieve more reliable long-term predictions, including time horizons of up to 5 seconds~\cite{Rondón2022}. Unlike methods that rely solely on past head positions, which perform well for short-term predictions such as 1 second~\cite{Qian2016} but degrade over longer periods, integrating saliency estimation offers a more effective and sustained approach to predict viewer focus over extended durations.
In this section, we review the scientific literature concerning saliency estimation both for 2D and 360-degree content.

%\subsection{2D content}
A variety of saliency estimation models have been proposed for traditional images and videos. For instance, in~\cite{Lou_2022} transformers and \acp{CNN} have been combined to predict saliency maps from static images. 

In the video domain, TASED-Net~\cite{Min2019} used a 3D convolutional encoder-decoder architecture. The encoder extracts spatial and temporal features from input frames, while the decoder reconstructs saliency maps based on the encoded representation. STSA-Net~\cite{Wang2021_2} introduced a spatio-temporal self-attention 3D network that incorporates multiple self-attention modules between 3D convolutional layers to capture long-range temporal dependencies between frames. Similarly,~\cite{Wang2021} and~\cite{Wu2020} proposed \ac{Conv-LSTM} frameworks that integrate \acp{CNN} with \ac{LSTM} units and attention mechanisms to model temporal dynamics more effectively. Furthermore, Droste \textit{et al.}~\cite{Droste2020} developed the UNISAL model, which utilizes the lightweight MobileNet~\cite{howard2017} architecture to facilitate efficient saliency estimation in both image and video domains.

%\subsection{360-degree content}

Many studies have been devoted to 360-degree saliency estimation for both images and videos. For static 360-degree images, some approaches employed feature-based methods such as~\cite{BATTISTI2018}~\cite{Baldoni_ISPA_2023}, while more advanced methods used deep learning. Nguyen \textit{et al.} proposed PanoSalNet~\cite{Nguyen2018}, which uses CNN layers specifically designed for 360-degree images. This method has been included in viewport prediction pipelines in many approaches, such as~\cite{Rondón2020}~\cite{Tian2025}. Similarly, SALGAN360~\cite{Chao2018} adapted the 2D saliency model SALGAN~\cite{Junting2018} by fine-tuning it with 360-degree data, using a \ac{GNN}~\cite{goodfellow2014} to predict saliency maps. In~\cite{martin2020} an encoder-decoder architecture with spherical convolutions has been employed. Spherical convolutions have been presented in~\cite{Coors2018} and consist of redefining the kernels to account for the distortions in the equirectangular projection, effectively modeling the geometry of omnidirectional images. In~\cite{Zhu_2020}, a saliency estimator for omnidirectional images has been proposed and integrated into a head movement prediction framework to reduce latency and bandwidth consumption. The saliency estimator is designed by fusing low and high level features, such as color information, car and person detection, equator bias, and other cues.

In 360-degree videos, models typically integrate spatial and temporal features. Incorporating the concept of spherical convolutions, Sphere-GAN~\cite{wahba2025} introduces a \ac{GAN}-based approach, where the generator consists of a U-Net~\cite{Olaf2015} that employs spherical convolutions, while the discriminator uses standard conventional \ac{CNN} layers. %Additionally, Sphere-GAN predicts the saliency map at time $t$ using the current equirectangular frame and the ground truth saliency map from time $t-5$.
As an alternative approach, CP360~\cite{cheng2018} projected the equirectangular frames in cubemap format and then applied \ac{CNN} layers combined with \ac{Conv-LSTM}~\cite{Xingjian2015} to capture spatio-temporal dependencies. \ac{RL} has also been introduced to estimate saliency maps offline while simultaneously predicting viewport trajectories online~\cite{Xu2018_2}. SphereU-Net~\cite{Zhang_ECCV_2018} extended the U-Net framework with an alternative version of spherical convolutions, where kernels are defined on spherical crowns and shifted across the sphere to respect the spherical geometry. A spherical 3D CNN-based %encoder–decoder 
model has been proposed in~\cite{Chen_2023}, where outputs were fused with center and initial frame biases for better prediction. 

Attention mechanisms have also been exploited. ATSal~\cite{Yasser2020} incorporated an attention module to encode global static spatial features in the equirectangular domain, while simultaneously using an expert module on local cubemap patches across frames, fusing the outputs for final saliency estimation. Another valuable approach is SPVP360~\cite{Jie2022}, which combines spherical \acp{CNN} with a dual encoder-decoder design, featuring one stream for spatial features and one for temporal ones. The output of the two branches are then fused through an attention mechanism. In~\cite{Wan2024}, another dual-stream framework used spherical \ac{Conv-LSTM} and attention modules to process forward and backward frame sequences, with Gaussian bias maps added during post-processing. In SVGC-AVA~\cite{Yang2024_2}, the authors proposed a model based on spherical vector-based graph convolutions and audio-visual attention, where visual and audio features were extracted in the spherical domain. %They incorporated the audio-visual attention mechanism within the U-Net framework. 

Another option consists in incorporating optical flow information. The authors of~\cite{Zhang2020}, propose a framework combining cubemap projections with optical flow, where spatial and temporal features were extracted using \acp{CNN} and refined via a bidirectional \ac{Conv-LSTM}-based saliency module. In addition, SST-Sal~\cite{Edurne2022} introduced an encoder-decoder model that integrates optical flow with spherical \ac{Conv-LSTM} layers to capture spatio-temporal dynamics. Optical flow is employed also in 360Spred~\cite{Yang2024}, where a U-Net architecture and a 3D separable graph convolutional network~\cite{Yang2020} are used. 360Spred takes spherical frames and spherical optical flows as inputs. These optical flows are computed directly in the spherical domain using a spherical graph-based approach. This framework effectively extracts both visual and motion features within the sphere domain and leverages temporal correlations across both high-level and low-level spatial features. In~\cite{Zhu_2022} and~\cite{Zhu_2023}, a graph-based model was employed for omnidirectional videos. Each omnidirectional frame was first partitioned into small blocks. A graph-based framework was then constructed to integrate the saliency information of these blocks with optical flow cues and the blocks themselves, enabling the generation of the final saliency maps.

%In contrast, we propose a lightweight model, SalFormer360. Our model is based on a transformer architecture specifically designed for 360-degree video saliency prediction. 
%It leverages a pretrained 2D Segformer encoder~\cite{Enze_2021} and integrates a custom 360-degree decoder with fixation biases. 
%Our model successfully estimates saliency maps for 360-degree frames using only 2 frames, outperforming existing state-of-the-art approaches as illustrated in Figure~\ref{fig:graphabs}.

Although many saliency estimation methods have tried to tackle saliency estimation for 360-degree videos, several limitations still exist.
Many approaches rely on spherical convolutions or on converting equirectangular frames to cubic projections to address geometric distortions, thus introducing computational overhead. 
Additional complexity arises from the use of optical flow for motion-aware saliency estimation or from the processing of long temporal sequences (\textit{e.g.}, more than $10$ frames~\cite{Chen_2023,Wan2024}).
Although these techniques can preserve temporal consistency, they significantly increase computational complexity and do not always lead to proportional improvements in performance.

In contrast to prior works that rely on computationally expensive solutions, we propose a lightweight transformer-based model, SalFormer360, that takes full advantage of the transformers’ powerful ability to capture long-range dependencies and global contextual relationships~\cite{Ashish2017} within 360-degree video content. Using this capability, our model efficiently learns both spatial and temporal correlations using only two frames (the current frame and the one at $t-5$). By integrating the viewing bias and a customized decoder, SalFormer360 outperforms state-of-the-art saliency prediction methods on 360-degree video benchmarks, confirming the %strong 
representational power of transformer architectures while maintaining low computational complexity. %, as illustrated in Figure~\ref{fig:graphabs}.  

\section{SalFormer360 Model}\label{sec:salformer}

%\section{Problem Formulation}

In 2D images and videos, the \ac{FoV} is fixed, making saliency estimation relatively straightforward. In contrast, in 360-degree videos, users can freely explore any direction within the scene, increasing the complexity of the saliency estimation and prediction. %and making saliency prediction far more challenging.
%To enable the processing and transmission of 360-degree content, many projection methods are used, one of the most common being the equirectangular projection. However, this projection introduces distortions, particularly near the poles, which further complicate the saliency estimation task.As a result, estimating saliency in 360-degree videos is considerably more challenging than in traditional 2D images and videos.

Given a sequence of 360-degree video frames $\mathbf{I} \in \mathbb{R}^{T \times C \times H \times W}$, where $T$ is the number of frames, $C$ is the number of channels, and $H \times W$ is the spatial resolution,  the objective is to predict the corresponding saliency maps $\mathbf{S} \in [0,1]^{T \times 1 \times H \times W}$ highlighting regions of interest, which are likely to attract user attention. 

\subsection{Network Structure}

Let $\mathbf{I}_t$ and $\mathbf{I}_{t-k}$ represent the input frames at time $t$ and $t-k$, respectively. The objective is to predict the saliency map $\mathbf{S}_t$ for frame $t$. This goal can be formulated as:

\begin{equation}
\mathbf{S}_t = \mathcal{M}(\mathbf{I}_t, \mathbf{I}_{t-k}),
\end{equation}
where $\mathcal{M}$ denotes the saliency estimation model that takes both the current and the previous frame as input.

%To tackle saliency estimation task, we utilize a transformer-based encoder, SegFormer~\cite{Enze_2021}, to capture detailed spatial representations. Additionally, we design a custom decoder tailored to our specific task, enabling the generation of saliency maps that are well-suited to our desired 360-degree frames.

\begin{figure}[t]
    \centering
    \includegraphics[width=.9\linewidth]{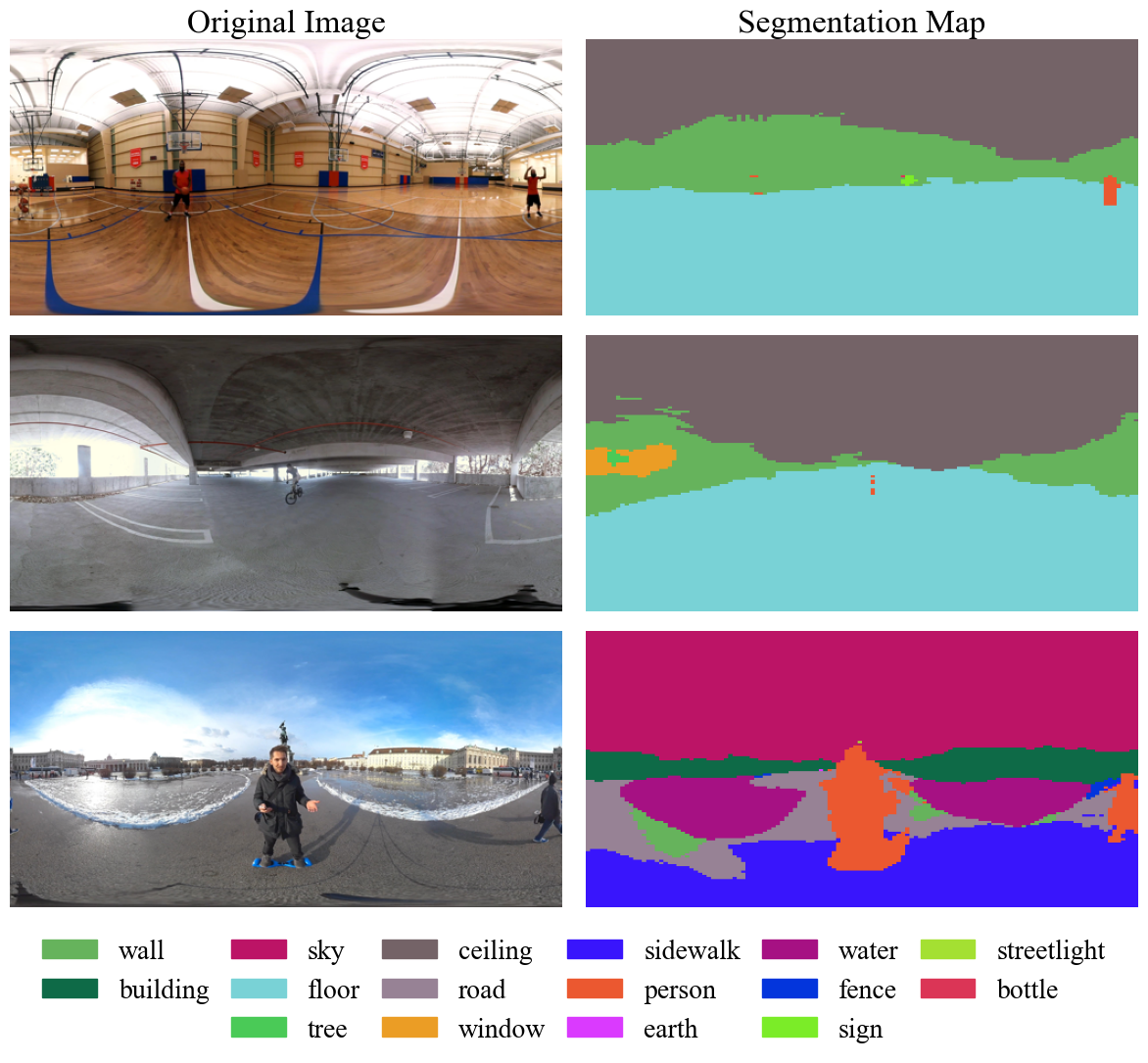}
    \caption{Segmentation results obtained by feeding 360-degree equirectangular frames into the SegFormer-B0 model.} 
    %\label{Segmentation results obtained by feeding 360-degree equirectangular frames into the SegFormer model (MiT-B0)}
    \label{fig:segmantation}
\end{figure}

To tackle the task of saliency estimation in 360-degree videos, we employ a transformer-based architecture that effectively captures both local and global spatial features. Specifically, we utilize the encoder of the SegFormer model~\cite{Enze_2021} due to its efficiency and ability to extract detailed spatial representations through multiscale feature learning. The encoder was fine‑tuned in conjunction with a custom decoder, designed to generate saliency maps from the 360-degree frame inputs.

As SegFormer was originally designed and trained for semantic segmentation on 2D images, we conducted an initial experiment to verify its suitability for 360-degree content. Specifically, our task involves processing 360-degree frames represented in equirectangular format. This projection introduces geometric distortions, especially near the poles, due to the transformation from spherical to planar coordinates. To assess the impact of geometric distortions on SegFormer's performance in segmentation tasks, we input the equirectangular frames into the lightweight SegFormer-B0 model, without applying any architectural modification, and qualitatively evaluate the output. 
This version of the segmenter was trained on \mbox{ImageNet-1K~\cite{Deng2009}} and fine-tuned on ADE20K~\cite{Bolei_2018}.

Figure~\ref{fig:segmantation} presents three examples of omnidirectional image segmentation obtained by using SegFormer. The model shows a strong ability to recognize meaningful visual content, successfully identifying objects such as people, sky, and floor, despite projection-related distortions. 
This observation validates the potential of using SegFormer as an effective feature extractor for our task.

\begin{figure*}[t]
    \centering
    %\graphicspath{{images/}}
    \includegraphics[width=0.95\linewidth]{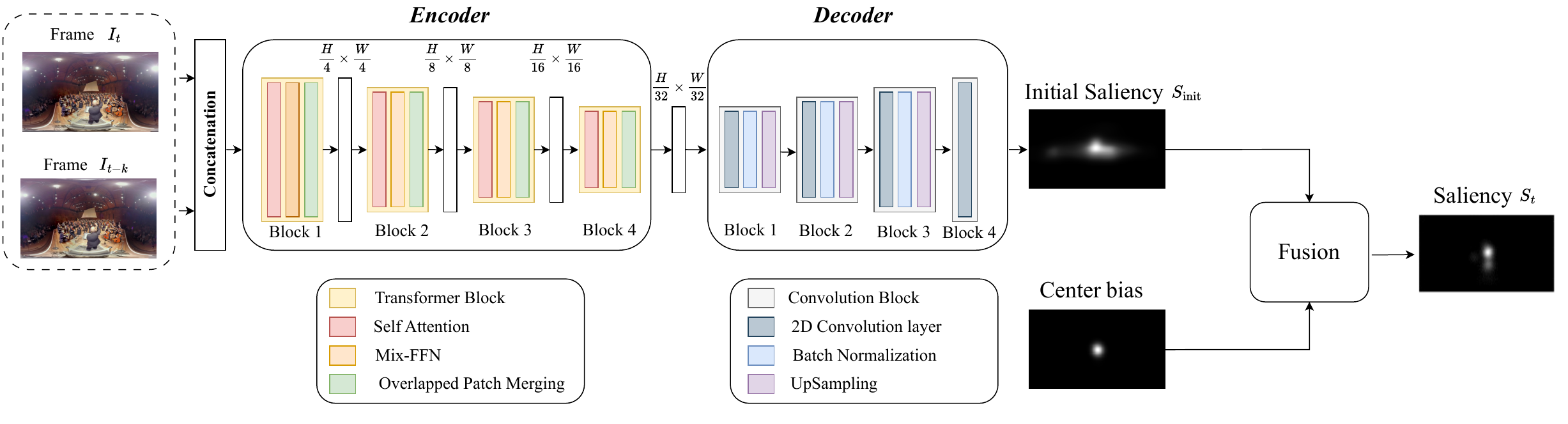}
    %\noindent\makebox[\textwidth]{\includegraphics[width=\paperwidth]{images/Structure.png}}
    \caption{Overview of the proposed 360-degree saliency estimation framework.} 
    \label{fig:dig}
\end{figure*}

Building on this foundation, we adopt the pre-trained and fine-tuned MiT-B0 encoder as the backbone of our model. To adapt it for the specific challenges of saliency estimation in 360-degree videos, we design a custom decoder that utilizes the features extracted by the encoder to produce an initial saliency map. This initial prediction is then adaptively fused with a \ac{CB} to generate the final saliency output. The viewing \ac{CB} accounts for the tendency of human observers to focus on the center of the 360-degree frame, as will be detailed in Section~\ref{CB}. Figure~\ref{fig:dig} shows the overall architecture of our proposed framework. In the following subsections, we provide a detailed explanation of each component.

\subsubsection{Encoder}\label{sec:encoder}
%For our model, we adopt the encoder component of SegFormer-B0 (MiT-B0), due to its efficiency in capturing both local and global features. SegFormer is a lightweight transformer-based architecture originally developed for semantic segmentation on 2D images. It combines the representational power of transformers with effective multi-scale feature learning. Specifically, we utilize the MiT-B0 encoder from Segformer-B0, which is the lightest version of SegFormer. This model is pre-trained on the ImageNet-1K dataset and subsequently fine-tuned on the ADE20K dataset. MiT-B0 encoder follows 
For our model, we adopt the encoder component of SegFormer-B0 (MiT-B0), which implements a hierarchical transformer architecture composed of multiple stages. Each stage includes a transformer block that reduces spatial resolution through overlapping patch embeddings and then applies self-attention and feedforward layers. 
While the original SegFormer encoder is designed to process a single RGB image (with $3$ channels), we modified it to handle two inputs, the current frames at time $t$ and the frame at time $t-k$. We concatenated the two frames along the channel dimension, resulting in a $6$-channel input. To this aim, we adapted the first convolutional layer of the encoder to accept $6$ channels by duplicating the pre-trained weights for the first $3$ channels.
After processing the input through the encoder, we obtain a set of feature maps that effectively capture the spatial structure and semantic content of the 360-degree video frames.

Differently from the works that rely on the optical flow, our method captures temporal dynamics implicitly. Indeed, the transformer-based architecture processes the concatenated frames to simultaneously capture the local spatial structure and temporal motion.
This approach avoids the overhead and instability often associated with optical flow computation while still capturing the temporal cues needed for 360-degree video saliency. Concerning the number of frames that are concatenated, we tested our approach with different frame counts ranging from $2$ to $10$. The obtained results highlighted that the increased computational cost due to the processing of multiple frames did not correspond to an increase in performance.

\subsubsection{Decoder}\label{subsec:decoder}
 %In our decoder, we employ spherical convolution instead of traditional 2D convolution. The concept of spherical convolution, introduced in SphereNet~\cite{Coors_2018}, was originally developed for tasks such as image classification and object detection for 360-degree content. The core idea is to adapt convolution and pooling operations from the conventional 2D image domain to the spherical domain.
%This is achieved by first applying the standard convolutional kernel to the tangent plane of the sphere. The kernel is then projected onto the equirectangular representation, where it undergoes the same type of distortion as the input equirectangular image. As the distorted kernel moves across the equirectangular image, it enables the network to effectively learn spatial features within the distorted content. Using spherical convolution offers two key advantages. First, it helps mitigate the distortion introduced by the equirectangular projection, preserving the geometric integrity of objects across the frame. Second, it addresses the discontinuities that often occur at image boundaries, where objects near the edges can appear split due to the transformation from spherical to planar representation. These benefits make spherical convolution particularly well-suited for saliency estimation in 360-degree video frames, where maintaining spatial consistency and geometric awareness is crucial.

The proposed decoder architecture is specifically tailored to the generation of 360-degree saliency maps. Our decoder takes as input the feature map provided by the last stage of the encoder, which has the lowest spatial resolution. 
The decoder progressively reduces the channel dimensionality while increasing the spatial resolution. The feature map is passed through a sequence of three convolution layers, each followed by batch normalization, a ReLU activation function, and an upsampling layer using bilinear interpolation.
The first convolutional layer maps the latent representation from $256$ to $128$ channels, and uses upsampling with a factor of $4$. The second layer further reduces the channels from $128$ to $64$ and applies the same upsampling factor. The third convolution reduces the number of channels from $64$ to $16$ with an upsampling factor of $2$. 
Finally, a convolutional layer maps the $16$ channel feature map to a single channel using a sigmoid activation function to produce the initial saliency map ($\mathbf{S}_{\text{init}}$). All convolutional layers use a kernel size of $3\times3$, stride $1$, and padding $1$. Furthermore, to improve generalization and prevent overfitting, during training we apply dropout with a rate of $30\%$ throughout the decoder.

\subsubsection{Viewing center bias}\label{subsubsec:CB}
\begin{figure}[t]
     \centering
     \subfloat[]{\includegraphics[width=0.32\linewidth]{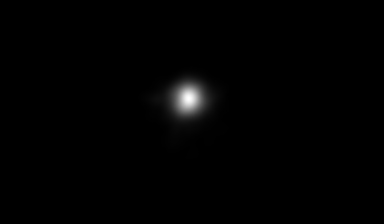}\label{fig:0314}}\hspace{0.2pt}
     \subfloat[]{\includegraphics[width=0.32\linewidth]{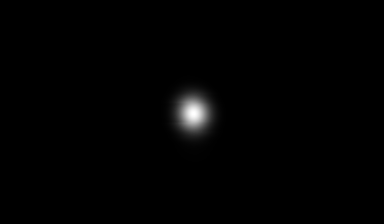}\label{fig:0314_gt}}\hspace{0.2pt}
     \subfloat[]{\includegraphics[width=0.32\linewidth]{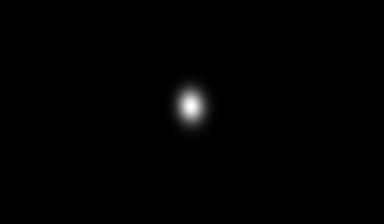}\label{fig:0314_ours}}\\
     
     \caption{Viewing Center Bias (CB),  refer to Sec.~\ref{subsubsec:CB} for details.
     (a)~PVS-HM dataset, (b)~Sport360 dataset, (c)~VR-EyeTracking dataset.}
     \label{CB}
\end{figure}

When users begin watching 360-degree videos, the headset typically starts by displaying the same default viewport, meaning that the same longitude and latitude coordinates are shown at the beginning for all users. The device configuration and setup determine this starting viewpoint. Inspired by the approach proposed in~\cite{Chen_2023}, we computed the average ground truth saliency map of the first frame in all (training) videos in our 3 datasets, resulting in a representative image referred to as the \ac{CB}, as illustrated in Figure~\ref{CB}. This \ac{CB} captures the common visual focus present at the beginning of the 360-degree video playback. Moreover, our experiments and previous literature~\cite{Chen_2023,Wan2024,Zhang2020}, show that this viewing bias is present throughout the video experience, making it an important component for saliency estimation models in 360-degree content. However, as the user is free to explore the content in all viewing direction, the \ac{CB} decreases over time~\cite{Chen_2023}. 
This means that the \ac{CB} contributes more to the estimation of the saliency of the initial frames, and contributes less as the frame number increases.
Therefore, we incorporated the \ac{CB} into our model using the combination of a fixed term and a decaying function. The latter is guided by learnable parameters, which are optimized during the training process. % (see Eq. \ref{eq:CB_frame}).
%\begin{figure}[htp]
%    \centering
    %\graphicspath{{images/}}
%    \includegraphics[width=7cm]{images/ifb.png}
%    \caption{Initial Frame Bias (IFB)} 
%    \label{360 video}
%\end{figure}

%Following the findings of~\cite{Chen_2023}, and based on the observation that users tend to explore the scene more freely after the initial moments, gradually shifting their gaze away from the starting view, we integrate the \ac{IFCB} into our saliency prediction framework with a dynamic trend. 
%More in detail, the \ac{CB} is fused with the initial prediction result of the model in an adaptive weight function, which can use a time decay function to dynamically learn the fusing weights of \ac{CB} map and the initial prediction result of the model~\cite{Chen_2023}. % to generate the last result. 

Formally, the estimation of the final saliency map ($\mathbf{S}$) with the biasing operation can be described as:
\begin{equation}
 \mathbf{S} = w_t \cdot \mathbf{CB} + (1 - w_t) \cdot \mathbf{S}_{\text{init}}, \label{eq:CB_frame}
\end{equation}
where %$\mathbf{S}_{IFCB}$ is the estimated saliency map for frame $t$,
$\mathbf{S}_{\text{init}}$ is the unbiased saliency prediction for frame $t$, and  
$w_t$ is a time-dependent weighting factor calculated as follows:
\begin{equation} \label{eq:CB_weight}
w_t = (1-\beta_i)\delta(t) + \beta_i.
\end{equation}
Here, $t$ denotes the current frame index, $\delta(t) = e^{-\alpha_i (t/C)^2}$ is the function guiding the dynamics of the biasing~\cite{Chen_2023}, $\alpha_{i}$ and $\beta_{i}$ are learnable parameters trained along with the model, where $i$ indicates the dataset used for training the model, and $C=600$ is a constant.

It is useful to note that, thanks to the separation between dynamic and static components of the weighting factor, our approach can effectively model long and short-range biasing concurrently. The short-term bias is modeled by the dynamic function $\delta(t)$, which encodes the fact that the center bias is more prominent in the initial frames. As the frame index increases, $\delta$ gradually decays towards zero, % (controlled by the parameter $\alpha$), 
mimicking the users' exploration behavior. %, and the model relies more on long-term bias. 
In some videos, a subtle bias towards the center of the scene remains, and this is addressed by the long-term bias, controlled by the static parameter $\beta_i$, which remains constant in time. Both bias components are adaptively controlled by learnable parameters ($\alpha_i$ and $\beta_i$), which regulate the strength of the short-term and long-term biases, respectively. These parameters, different for each dataset, are %automatically 
optimized during training%and vary depending on the characteristics of each dataset
, allowing the model to adjust its biasing behavior to the specific viewing tendencies present in the data.

\subsection{Loss Function}
To train our model, we used a loss function as a combination of four different losses: \ac{CC} loss, \(\mathcal{L}_{\text{CC}}\), \ac{KL} loss, \(\mathcal{L}_{\text{KL}}\), \ac{SMSE} loss, \(\mathcal{L}_{\text{SMSE}}\), and \ac{BCE} loss, \(\mathcal{L}_{\text{BCE}}\). The final loss is illustrated in Equation~\ref{eq:1}:
\begin{equation}
\label{eq:1}
\begin{split}
\mathcal{L} =& \mathcal{L}_{\text{CC}}(\hat{\mathbf{S}}, \mathbf{S}) + \mathcal{L}_{\text{KL}}(\hat{\mathbf{S}}, \mathbf{S}) + \\ &\mathcal{L}_{\text{SMSE}}(\hat{\mathbf{S}}, \mathbf{S}) + \mathcal{L}_{\text{BCE}}(1, \mathbf{S}), 
\end{split}
\end{equation}
where $\hat{\mathbf{S}}$ denotes the ground truth saliency map, while $\mathbf{S}$ represents the predicted saliency map. The \ac{CC} loss is defined as $\mathcal{L}_{\text{CC}}(\hat{\mathbf{S}}, \mathbf{S}) = 1 - \text{\( CC \)}(\hat{\mathbf{S}}, \mathbf{S})$, where $CC$ is the Pearson linear correlation coefficient that measures the linear correlation between the predicted and ground truth saliency maps. The \ac{KL} loss is defined as $\mathcal{L}_{\text{KL}}(\hat{\mathbf{S}}, \mathbf{S}) = \text{\( KL \)}(\hat{\mathbf{S}}, \mathbf{S})$, which quantifies the difference between the predicted saliency distribution and the ground truth. The \ac{SMSE} loss is defined as  $L_{\text{SMSE}}(\hat{\mathbf{S}}, \mathbf{S}) = \Psi(\theta, \phi) \times MSE(\hat{\mathbf{S}}, \mathbf{S})$ where $MSE$ represents the mean squared error, and $\Psi(\theta, \phi)$ represents the spherical weights that allow giving more importance to regions near the equator and lower weights to areas located near the poles. The $L_{\text{SMSE}}$ accounts for geometric distortions in equirectangular projections by penalizing prediction errors while respecting spherical geometry. Finally, the \ac{BCE} loss, $\mathcal{L}_{\text{BCE}}$, is a pixel-wise loss function that compares the predicted saliency map with the ground truth, focusing on accurately locating salient points.

\section{Experimental results}\label{sec:results}

\subsection{Datasets}\label{subsec:dataset}
%To train and test our model, we used the dataset proposed by~\cite{Zhang_ECCV_2018}, which consists of 104 360-degree videos with durations ranging from 20 to 60 seconds and features five different types of sports. This dataset was created by having 20 users view videos using the HTC VIVE equipped with the 7invensun a-Glass eye tracker. It includes colored frames, ground truth saliency maps, and the original gaze points of the viewers. For our model, we selected 80 videos for training and 24 videos for testing, following the original splitting method outlined by the authors.

To train and test our model, we used the three largest open-source datasets %commonly used 
for saliency estimation and viewport prediction: PVS-HM~\cite{Xu2018_2}, Sport360~\cite{Zhang_ECCV_2018}, and VR-EyeTracking~\cite{Xu2018}.

The PVS-HM dataset consists of 76 360-degree videos, with durations varying from 20 to 80 seconds, with different resolutions from 3K to 8K. These videos were viewed by 58 participants using the HTC VIVE \ac{HMD}. The dataset includes various content types, such as gaming, sports, animation, driving, and natural scenes. 
The dataset contains MP4 videos and the corresponding head orientations of the viewing users.

The Sport360 dataset contains 104 360-degree videos, with durations ranging from 20 to 60 seconds, focusing on five different types of sports. This dataset was created thanks to the participation of 20 users who viewed the videos using the HTC VIVE equipped with the 7invensun a-Glass eye tracker. The dataset provides the RGB frames, the ground truth saliency maps, and the original gaze points of the viewers.

The VR-EyeTracking dataset consists of 208 360-degree videos having varying durations (from 20 to 60 seconds), covering a wide range of content, including documentaries, sports, and various indoor and outdoor scenes. Each video was watched by 45 participants using the HTC VIVE headset paired with the 7invensun a-Glass eye tracker. The dataset contains MP4 videos along with the eye fixation data and head orientations of the users.

We divided each dataset into training and testing sets. For the PVS-HM dataset, 61 videos were used for training, while 15 were set aside for testing. The Sport360 dataset was split into 80 videos for training and 24 for testing. Lastly, the VR-EyeTracking dataset was divided into 134 videos for training and 74 for testing. This splitting process followed the original methods outlined by the authors of each dataset.

% \subsubsection{Saliency maps generation}
\noindent\textbf{Saliency maps generation:}
To customize the PVS-HM and VR-EyeTracking datasets for our task, we generated RGB frames from the MP4 videos using FFmpeg 7.0.1\footnote{\url{https://www.ffmpeg.org/}} and then generated saliency maps for every dataset from the provided head-orientation data.
For the PVS-HM dataset, saliency maps were generated directly from user head-orientation recordings provided as longitude–latitude fixation points. Since the fixation data were sampled at twice the frame rate of the video (for example, 60 Hz vs. 30 Hz), we downsampled the fixation traces to match the video frame rate. For each frame, fixations were projected onto a 512×256 pixels equirectangular grid, and a Gaussian kernel with a standard deviation of 7 degrees was applied around each fixation to highlight fixation locations and produce smooth, continuous saliency maps. The contributions of all users were accumulated, and the resulting saliency maps were normalized to the range [0,1]. This process yielded one saliency map per frame, aligned with the video timeline.
For the VR-EyeTracking dataset, gaze fixations were provided as normalized pixel coordinates in the range [0,1]. These fixations were first scaled to the target equirectangular resolution ($512\times256$ pixels) and then converted into Gaussian-based saliency maps, where the Gaussian spread was computed in pixel units in order to correspond to a viewing angle of 7 degrees. To mitigate projection distortion, the intermediate saliency maps were projected into a cubemap representation, smoothed with a Gaussian filter on each cube face, and finally re-projected back into equirectangular format.

While both datasets were processed to produce $512\times256$ pixels saliency maps using a Gaussian kernel of size 7 degrees, the specific processing pipeline depended on the head orientation format in the original datasets. The PVS-HM dataset provided the head orientations in angular coordinates, which were directly mapped onto the sphere. In contrast, the VR-EyeTracking dataset provided the head orientation information in pixel-normalized coordinates, thus requiring cubemap smoothing to better handle projection distortions. The kernel size of 7 degrees was chosen to maintain consistency across datasets and to closely approximate the kernel used in the Sports-360 dataset (3.34 degrees for $128\times256$ pixels), ensuring comparable saliency spread across different spatial scales.

\begin{table*}[t]
\setlength{\tabcolsep}{.4em}
\centering
\caption{Performance comparison across Sport360~\cite{Zhang_ECCV_2018}, PVS-HM~\cite{Xu2018_2}, and VR-EyeTracking~\cite{Xu2018} datasets. Best results are in \textbf{bold}, second-best are \underline{underlined}.}
\begin{tabular}{
>{\centering\arraybackslash}p{0.04\textwidth}
!{\vrule width 0.4pt}>{\raggedright\arraybackslash}p{0.18\textwidth}!{\vrule width 0.4pt}
c:c:c:c!{\vrule width 0.4pt}
c:c:c:c!{\vrule width 0.4pt}
c:c:c:c
}
\toprule
\textbf{} & \textbf{Method} &
\multicolumn{4}{c!{\vrule width 0.4pt}}{\textbf{Sport360}} &
\multicolumn{4}{c!{\vrule width 0.4pt}}{\textbf{PVS-HM}} &
\multicolumn{4}{c}{\textbf{VR-EyeTracking}} \\
 & & CC ↑ & NSS ↑ & KL ↓ & AUC ↑ & CC ↑ & NSS ↑ & KL ↓ & AUC ↑ & CC ↑ & NSS ↑ & KL ↓ & AUC ↑ \\ \midrule

% ---------------- 2D Video ----------------
\multicolumn{1}{c!{\vrule width 0.4pt}}{\multirow{2}{*}{\shortstack{2D\\videos}}}
& TASED Net$^*$~\cite{Min2019}  & 0.352 & 1.912 & -- & 0.883 & 0.651 & 2.413 & -- & 0.905 & 0.201 & 2.009 & -- & 0.864 \\
\multicolumn{1}{c!{\vrule width 0.4pt}}{}
& STSA Net$^*$~\cite{Wang2021}  & 0.365 & 1.972 & -- & 0.931 & 0.476 & 1.710 & -- & 0.881 & 0.151 & 1.502 & -- & 0.869 \\ \midrule

% ---------------- 360 Image ----------------
\multicolumn{1}{c!{\vrule width 0.4pt}}{\multirow{3}{*}{\shortstack{360-degree\\images}}}
& PanoSalNet$^\ddagger$~\cite{Nguyen2018} & 0.489 & 2.981 & 13.344 & 0.633 & -- & -- & -- & -- & -- & -- & -- & -- \\
\multicolumn{1}{c!{\vrule width 0.4pt}}{}
& SalGAN360$^*$~\cite{Chao2018}  & 0.331 & 1.675 & -- & 0.859 & 0.443 & 1.603 & -- & -- & 0.236 & 1.267 & -- & 0.704 \\
\multicolumn{1}{c!{\vrule width 0.4pt}}{}
& Martin$^*$~\cite{martin2020}  & 0.287 & 1.496 & -- & 0.851 & 0.215 & 0.828 & -- & 0.736 & 0.225 & 2.322 & -- & 0.865 \\ \midrule

% ---------------- 360 Video ----------------
\multicolumn{1}{c!{\vrule width 0.4pt}}{\multirow{13}{*}{\shortstack{360-degree\\videos}}}
& CP360$^*$~\cite{cheng2018} & 0.243 & 1.007 & -- & 0.840 & 0.338 & 0.673 & -- & 0.758 & 0.104 & 0.973 & -- & 0.800 \\
\multicolumn{1}{c!{\vrule width 0.4pt}}{}
& Offline DHP$^*$~\cite{Xu2018_2}  & 0.445 & 2.591 & -- & 0.874 & 0.704 & 3.137 & -- & -- & -- & -- & -- & -- \\
\multicolumn{1}{c!{\vrule width 0.4pt}}{}
& Spherical U-Net$^*$~\cite{Zhang_ECCV_2018}  & 0.625 & 3.534 & -- & 0.898 & 0.745 & 3.175 & -- & -- & -- & -- & -- & -- \\
\multicolumn{1}{c!{\vrule width 0.4pt}}{}
& Zhang360$^*$~\cite{Zhang2020}  & 0.620 & \textbf{5.125} & -- & 0.937 & 0.767 & 3.289 & -- & -- & -- & -- & -- & -- \\
\multicolumn{1}{c!{\vrule width 0.4pt}}{}
& ATSal$^*$~\cite{Yasser2020}  & -- & -- & -- & -- & 0.564 & 2.489 & -- & 0.914 & 0.226 & 2.464 & -- & 0.869 \\
\multicolumn{1}{c!{\vrule width 0.4pt}}{}
& SST-Sal$^*$~\cite{Edurne2022}  & 0.439 & -- & 8.610 & -- & 0.424 & 1.244 & -- & 0.833 & \underline{0.500} & -- & \underline{7.371} & -- \\
\multicolumn{1}{c!{\vrule width 0.4pt}}{}
& SPVP360$^*$~\cite{Jie2022} & 0.623 & 1.174 & -- & 0.662 & -- & -- & -- & -- & -- & -- & -- & -- \\
\multicolumn{1}{c!{\vrule width 0.4pt}}{}
& 3DSphereNet$^*$~\cite{Chen_2023} & \underline{0.666} & 4.586 & \underline{4.853} & \underline{0.940} & 0.768 & 3.750 & \underline{3.208} & 0.933 & -- & -- & -- & -- \\
\multicolumn{1}{c!{\vrule width 0.4pt}}{}
& SVGC-AVA$^*$~\cite{Yang2024_2} & -- & -- & -- & -- & 0.734 & 3.554 & -- & -- & -- & -- & -- & -- \\
\multicolumn{1}{c!{\vrule width 0.4pt}}{}
& Pred360$^*$~\cite{Wan2024} & 0.656 & 4.381 & -- & 0.931 & 0.773 & 3.734 & -- & \underline{0.934} & 0.291 & \textbf{3.768} & -- & \underline{0.899} \\
\multicolumn{1}{c!{\vrule width 0.4pt}}{}
& 360Spred$^\dagger$~\cite{Yang2024} & 0.661 & -- & -- & 0.937 & \underline{0.787} & \underline{3.838} & -- & -- & -- & -- & -- & -- \\
% \cdashline{2-14}
\multicolumn{1}{c!{\vrule width 0.4pt}}{}
& \textbf{Ours}$^*$& \textbf{0.722} & \underline{4.746} & \textbf{3.267} & \textbf{0.943} & \textbf{0.807} & \textbf{3.946} & \textbf{1.042} & \textbf{0.937} & \textbf{0.593} & \underline{3.016} & \textbf{2.334} & \textbf{0.912} \\
\bottomrule
\end{tabular}
\label{table:results}
\end{table*}

\subsection{Experimental Setup}\label{sec:experimental_setup}
%To train our model, we used Adam optimizer and we implemented a custom learning rate schedule that combines a warm-up phase with exponential decay. During the first 10 epochs, we used a fixed learning rate of 0.0001 for the encoder, 0.001 for the decoder, and 0.1 for alpha, allowing the model to stabilize in early training.  After the warm-up phase, the learning rate decays exponentially at each epoch with a decay factor equal to 0.9. We also used a weight decay of 1e-5 for both the encoder and decoder, and a dropout rate of 30\% in the decoder side as a regularization term and to prevent overfitting for our model. The model was trained for 20 epochs with an early stopping strategy and utilized a batch size of 16 frames. Alpha is set as a learnable parameter with initial value equal to 1200, which had been chosen empirically. All implementations were completed using the PyTorch library, and the model was trained and tested using an NVIDIA L40S GPU.

%We resized our model to 224x384, inspired by~\cite{Chen_2023}. To help the model to focus on understanding the content of the frames rather than memorizing their locations, we applied horizontal and vertical random flipping with a probability of 50\%. Also we perforemed color jittering with random adjustments for brightness, contrast, and saturation, ranging from 0.7 to 1.3, to enhance the model's robustness against color variations. For training purposes, we set the sampling interval to 5, which means one frame was taken every 5 frames to prevent overfitting in our model.

For model training and testing, we used the PyTorch library~\cite{Adam2019} and an NVIDIA L40s GPU. We employed the Adam optimizer with a learning rate of $10^{-4}$ for the encoder and $10^{-3}$ for the decoder. %Empirically, we set the $\alpha$ learning rate to 0.1 and the $\beta$ learning rate to 0.0001, based on observed performance improvements during preliminary experiments. 
To incorporate a regularization term and prevent overfitting, we applied a weight decay of $10^{-4}$ for both the encoder and decoder, along with a $30\%$ dropout rate on the decoder side. The batch size for training and testing was set to $4$.  %and we trained our model for 14,600 iterations on the Sport360 dataset, 1,100 iterations for PVS-HM dataset, and 6,300 iterations on the VR-EyeTracking dataset. 
Thanks to the usage of a pretrained encoder, our model converges very fast, requiring only 6 epochs with early stopping (average of 3000 iterations to convergence).
%for training on the Sport360 dataset, and just one epoch for the PVS-HM and VR-EyeTracking datasets with early stopping. 
The Sport360 dataset has 2,891 iterations per epoch, while PVS-HM and VR-EyeTracking comprise 2,182 and 6,871 iterations per epoch, respectively. 
%One reason the PVS-HM dataset converges faster is that it is more center-biased compared to the other datasets, which helps the model to converge quickly. The VR-Eyetracking dataset converges within a single epoch because it requires more iterations per epoch (6,871) than both Sport360 (2,891) and PVS-HM (2,182), contributing to a longer training cycle. 
To avoid overfitting for the three datasets, we sampled the training set to include one every five frames. We also applied various data augmentation strategies to the dataset, such as horizontal and vertical flipping with $50\%$ and $5\%$ rates, respectively, to enhance the model learning capability. Furthermore, we implemented color jittering with random adjustments for brightness, contrast, and saturation, ranging from 0.7 to 1.3, to improve the model robustness against color variations~\cite{torchvision2016}. We resized the 360 frames to $224\times384$ pixels, inspired by~\cite{Chen_2023}. The parameters $\alpha$ and $\beta$ are defined as learnable variables, with initial values empirically set to 600 and 0.15, respectively. Their learning rates were also determined through preliminary experiments, with $\alpha$ assigned a learning rate of 0.1 and $\beta$ set to $10^{-4}$, based on observed performance improvements. Concerning the parameter $k$, we performed an empirical study, with values ranging from $1$ to $20$, and selected $k=5$ as it was the value which yielded the best performance.% and C is set to 600 as in~\cite{Chen_2023}.

%The generated saliency maps for PVS-HM and Eye-tracking, along with the corresponding RGB frames and the full processing code, will be made available on GitHub to support reproducibility and future research.

\subsection{Baseline methods}
To evaluate our model, we compare it against a wide range of baseline methods. These include 2D saliency prediction approaches, such as TASED Net~\cite{Min2019} and STSA Net~\cite{Wang2021}; 360-degree image saliency models, including PanoSalNet~\cite{Nguyen2018}, SalGAN360~\cite{Chao2018}, and the method proposed by Martin \textit{et al.}~\cite{martin2020}; as well as video saliency prediction techniques, such as CP360~\cite{cheng2018}, Offline-DHP~\cite{Xu2018_2}, SphericalU-Net~\cite{Zhang_ECCV_2018}, Zhang360~\cite{Zhang2020}, ATSal~\cite{Yasser2020}, SST-Sal~\cite{Edurne2022}, SPVP360~\cite{Jie2022}, 3DSphereNet~\cite{Chen_2023}, SVGC-AVA~\cite{Yang2024_2}, Pred360~\cite{Wan2024}, and 360Spred~\cite{Yang2024}.
The numerical results for %these methods 
were taken directly from the %corresponding 
original papers, with the exception of SalGAN360, SphericalU-Net, and Offline-DHP, whose results were obtained from~\cite{Zhang2020}. Similarly, results for CP360, TASED Net, STSA Net, and Martin \textit{et al.} were referenced from~\cite{Wan2024}, while those for PanoSalNet were taken from~\cite{Peng2023}.

%We summarize and describe some of these baselines below.
%\begin{itemize}
%\item PanoSalNet~\cite{Nguyen2018}: one of the famous saliency estimation methods for 360-degree images. The model consists of nine CNN layers, the first three layers initialized using VGG-Net~\cite{Simonyan2014} parameters. Transfer learning was applied by replacing the fully connected layers with new ones suitable for the saliency detection task.
%\item SST-Sal~\cite{Edurne2022}: this saliency estimation model is designed for 360-degree videos, where the U-Net framework combined with ConvLSTM has been used,incorporating optical flow to predict saliency maps.
%\item SPVP~\cite{Li_2023}: this model estimates the saliency for 360-degree videos by extracting spatial and temporal features using two U-Net frameworks and using a CBAM attention mechanism~\cite{Sanghyun2018} to predict saliency maps.
%\item 3DSphereNet~\cite{Chen_2023}: this saliency estimation model for 360-degree videos where a U-Net framework had been used with a 3D encoder-decoder architecture based on spherical convolution to estimate saliency maps.
%\item 360Spred~\cite{Yang2024}: this model used the U-Net structure along with a spherical graph-based algorithm and optical flow to estimate saliency maps for 360-degree videos.
%\item Pred360~\cite{Wan2024}: this saliency estimation model for 360-degree videos where a spherical ConvLSTM encoder-decoder model had been used to generate saliency maps.
%\end{itemize}

\subsection{Metrics}
To evaluate our model, we employed four metrics: \ac{KL}, \ac{CC}, \ac{NSS}, and \ac{AUC-Judd}. The \ac{CC}, and \ac{KL} were calculated between the ground truth and the predicted saliency maps. \ac{NSS} and \ac{AUC-Judd} were assessed between a binary map generated from the original fixation points and the produced saliency map, as both metrics are inherently defined with respect to human fixations. More details on these metrics can be found in~\cite{Bylinskii2019}. The metrics were implemented in the same way as the \textit{Salient360!} benchmark~\cite{Guti2018}. To tackle distortion issues during evaluation, particularly near the poles, we adopted a latitudinal sinusoidal factor for the saliency maps during evaluation. This approach involves assigning greater weights to salient points near the equator and smaller weights to those close to the poles~\cite{Guti2018}. For a fair comparison with other state-of-the-art approaches, our results in Table~\ref{table:results} adopt the following notation scheme. Methods that employed the tool with a latitudinal sinusoidal factor (default option) are marked with an asterisk ($*$). A dagger ($\dagger$) denotes methods that used the tool with uniform sampling, where evaluation metrics were calculated at uniformly sampled points on the sphere. A double dagger ($\ddagger$) indicates cases in which the procedure was undefined.

\begin{figure}[t]
    \centering
    % -------- 
    % \includegraphics[width=0.3\linewidth]{images/sport360/Test_vid306_fid318_original_image.pdf}\hspace{0.2pt}
    % \includegraphics[width=0.3\linewidth]{images/sport360/Test_vid306_fid318_ground_truth_saliency.png}
    % \includegraphics[width=0.3\linewidth]{images/sport360/Test_vid306_fid318_generated_saliency.png}\hspace{0.2pt}\\
    
    % -------- 
    %\vspace{4pt}
    %\includegraphics[width=0.3\linewidth]{images/sport360/Test_vid248_fid179_original_image.png}\hspace{0.2pt}
    %\includegraphics[width=0.3\linewidth]{images/sport360/Test_vid248_fid179_ground_truth_saliency.png}\hspace{0.2pt}
    %\includegraphics[width=0.3\linewidth]{images/sport360/Test_vid248_fid179_generated_saliency.png}\hspace{5pt}\\
   
    % \vspace{4pt}
    \includegraphics[width=0.48\linewidth]{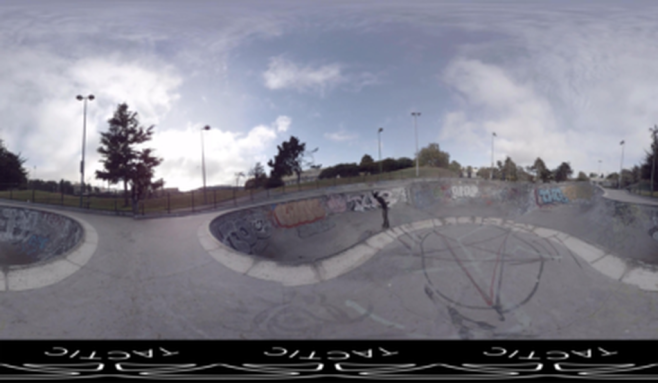}\hspace{0.2pt}
    \includegraphics[width=0.48\linewidth]{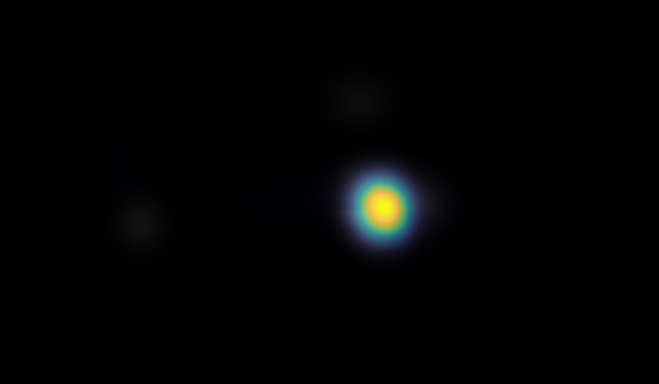}\hspace{0.2pt}
    \vspace{4pt}
    %\subfloat[]{
    \includegraphics[width=0.48\linewidth]{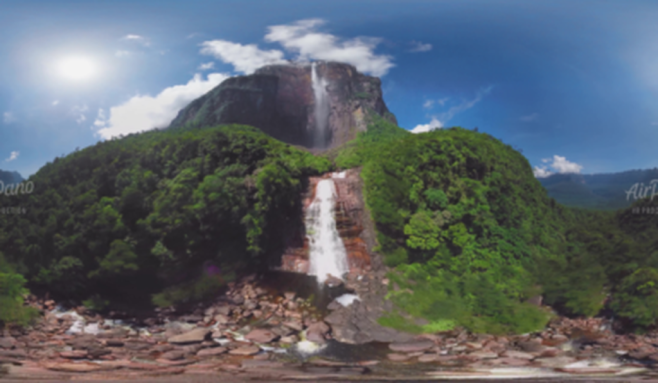}%}
    \hspace{0.2pt}
    %\subfloat[]{
    \includegraphics[width=0.48\linewidth]{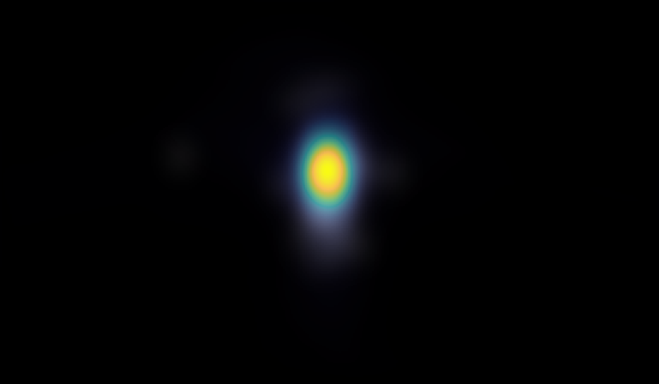}%}
    \hspace{0.2pt}
    %\subfloat[]{
    %\includegraphics[width=0.3\linewidth]{images/lowres/PVS/Test_vidWaterfall_fid66_generated_saliency.png}%}
    
%     \caption{
%         Qualitative results on \textbf{PVS-HM}: (left) three random images extracted from the adopted dataset, (center) ground-truth saliency
% maps, (right) saliency maps predicted with the proposed method.
%     }
%     \label{res-PVS-HM}
% \end{figure*}

% \begin{figure*}[ht!]
%     \centering
    % -------- 
    \vspace{4pt}
    \includegraphics[width=0.48\linewidth]{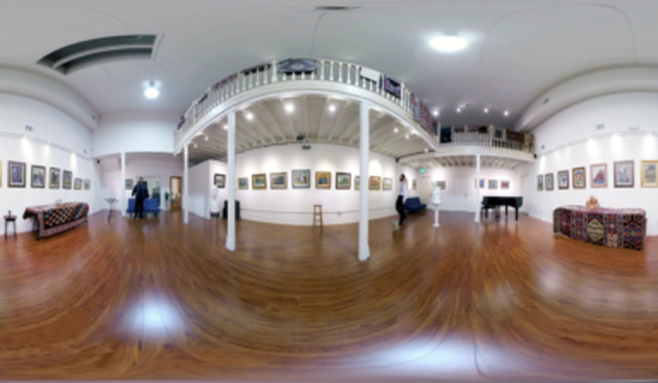}\hspace{0.2pt}
    \includegraphics[width=0.48\linewidth]{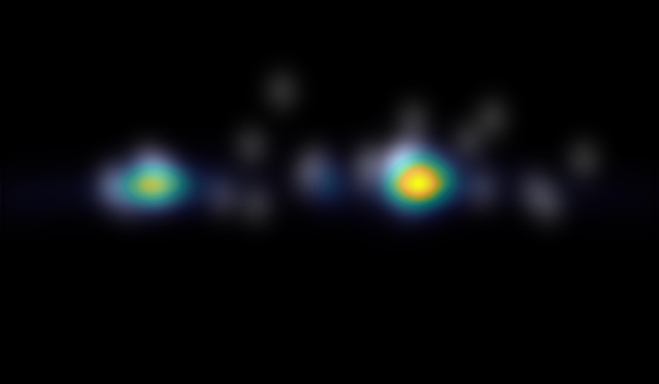}
    \hspace{0.2pt}\\
    
    % % -------- 
    % \vspace{4pt}
    % \includegraphics[width=0.3\linewidth]{images/eye/Test_vid143_fid71_original_image.pdf}\hspace{0.2pt}
    % \includegraphics[width=0.3\linewidth]{images/eye/Test_vid143_fid71_ground_truth_saliency.png}\hspace{0.2pt}
    % \includegraphics[width=0.3\linewidth]{images/eye/Test_vid143_fid71_generated_saliency.png}\hspace{5pt}\\
   
    %\vspace{4pt}
    %\includegraphics[width=0.32\linewidth]{images/eye/Test_vid177_fid222_original_image.png}\hspace{0.2pt}
    %\includegraphics[width=0.32\linewidth]{images/eye/Test_vid177_fid222_ground_truth_saliency.png}\hspace{0.2pt}
    %\includegraphics[width=0.32\linewidth]{images/eye/Test_vid177_fid222_generated_saliency.png}\\
   
    % -------- 
    % \vspace{-6pt}
    % \subfloat[]{\includegraphics[width=0.3\linewidth]{images/eye/Test_vid136_fid550_original_image.pdf}}\hspace{0.2pt}
    % \subfloat[]{\includegraphics[width=0.3\linewidth]{images/eye/Test_vid136_fid550_ground_truth_saliency.png}}\hspace{0.2pt}
    % \subfloat[]{\includegraphics[width=0.3\linewidth]{images/eye/Test_vid136_fid550_generated_saliency.png}}
    
    \caption{
        Qualitative results. First row: Sport360; second row: PVS-HM; third row: VR-EyeTracking. Left: RGB input; right: ground truth (grayscale) with an overlay of the estimated saliency map (parula colormap). %on \textbf{VR-EyeTracking}: (left) three random images extracted from the adopted dataset, (center) ground-truth saliency
%maps, (right) saliency maps predicted with the proposed method.
    }
    \label{fig:quali}
\end{figure}

\subsection{Performance Evaluation}
In this section we provide both a quantitative and qualitative performance assessment of the proposed method.

\subsubsection{Quantitative results}
The performance of our model is summarized in Table~\ref{table:results}, which shows that our approach consistently outperforms all state‑of‑the‑art methods across the three benchmark datasets in the categories of 2D videos, 360-degree images, and 360-degree videos.

In the 2D video category, the proposed model achieves substantial gains, with \ac{CC} improvements of 97.8\%, 24\%, and 195\% compared to the best prior methods on the Sport360, PVS-HM, and VR-EyeTracking datasets, respectively.

In the 360-degree image category, our model clearly outperforms image-based approaches like PanoSalNet and SalGAN360. In terms of CC, it achieves improvements of 47.6\%, 82.2\%, and 151.3\% on Sport360, PVS-HM, and VR-EyeTracking, respectively, compared to the best-performing 360-degree image-based methods. These results highlight the effectiveness of using temporal cues in our model design.

Finally, in the 360-degree video category, our model delivers the strongest overall performance with improvements of 8.4\%, 2.5\%, and 18.6\% in \ac{CC} on the Sport360, PVS-HM, and VR-EyeTracking datasets, respectively, compared to prior works. The only exceptions are noted in the \ac{NSS} metric, where our model ranks second behind Zhang360 on the Sport360 dataset and behind Pred360 on VR-EyeTracking.

%On the Sport360 dataset, our model achieves the best performance in three metrics, with a \ac{CC} of 0.722, \ac{KL} of 3.267, and \ac{AUC-Judd} of 0.943, while obtaining the second-best results in terms of \ac{NSS} with a score of 4.746. These results clearly show that the proposed method outperforms the state of the art approaches 
%the best recent approaches such as in 3DSphereNet (\ac{CC} = 0.666, \ac{NSS} = 4.586, \ac{KL} = 4.853, \ac{AUC-Judd} = 0.940), Pred360 (\ac{CC} = 0.6557, \ac{NSS} = 4.3808, \ac{AUC-Judd} = 0.9311), 360Spred (\ac{CC} = 0.6610, \ac{AUC-Judd} = 0.9370), and Pred360 (\ac{CC} = 0.656, \ac{NSS} = 4.381, \ac{AUC-Judd} = 0.931) which demonstrates the robustness of our framework. 

%On the PVS-HM dataset, our method achieved superior performance against other state-of-the-art approaches in terms of all the analyzed metrics, with \ac{CC} = 0.807, \ac{NSS} = 3.946, \ac{KL} = 1.042, and \ac{AUC-Judd} = 0.937.  It outperforms the best state-of-the-art approaches like 360Spred (\ac{CC} = 0.787, \ac{NSS} = 3.838), Pred360 (\ac{CC} = 0.773, \ac{NSS} = 3.734, \ac{AUC-Judd} = 0.934),  and 3DSphereNet (\ac{CC} = 0.768, \ac{NSS} = 3.750, \ac{KL} = 3.208, \ac{AUC-Judd} = 0.933). 

%On the VR-EyeTracking dataset, our approach has the best overall performance in terms of correlation (\ac{CC} = 0.593), \ac{KL} (\ac{KL} = 2.334), and \ac{AUC-Judd} (0.912), and achieves competitive results on NSS (3.016). Only the Pred360 method achieves a slightly higher \ac{NSS} of 3.768, while having worse performance in terms of other metrics. 

These results demonstrate that our model consistently improves saliency prediction accuracy across diverse datasets compared to several recent state-of-the-art approaches.

\subsubsection{Qualitative results}
To visually validate the effectiveness of our model, we present three randomly selected samples from the three datasets in Figure~\ref{fig:quali}. The left column shows the input RGB video frame, and the right column presents the corresponding grayscale ground-truth saliency map with an overlay of the estimated saliency map (parula colormap). %More precisely, Figure~\ref{res-Sport360} corresponds to the Sport360 dataset, Figure~\ref{res-PVS-HM} to the PVS-HM dataset, and Figure~\ref{res-VR-Eyetracking} to the VR-EyeTracking dataset. 
As clearly shown by the figure, our model successfully predicts accurate saliency maps across different datasets, providing clear evidence of its effectiveness.

\subsubsection{Model complexity and computational efficiency}

% \mycomment{
% \begin{table}[t]
% \centering
% \caption{Comparison of model complexity based on number of parameters.}
% \label{tab:params_final}
% \begin{tabular}{p{0.35\textwidth}|c}
% \toprule
% \multicolumn{2}{c}{\textbf{Number of Parameters}} \\ \midrule
% \multicolumn{1}{c|}{Methods} & Parameters (M) ↓ \\ \midrule
% Offline\_DHP (2018) & 34.00 \\ \hline
% SalGAN360 (2018) & 31.78 \\ \hline
% Spherical U-Net (2018) & 12.30 \\ \hline
% SST\_Sal (2022) & 8.32 \\ \hline
% SPVP360 (2023) & 7.72 \\ \hline
% 360Spred (2024) & 7.59 \\ \hline
% Ours & \textbf{3.70} \\  \bottomrule
% \end{tabular}
% \vspace{-1em}
% \end{table}

% \begin{table}[t]
% \centering
% \caption{Comparison of computational efficiency across models.}
% \label{tab:efficiency}
% \begin{tabular}{p{0.2\textwidth}|c|c}
% \toprule
% \multicolumn{3}{c}{\textbf{Model Efficiency}} \\ \midrule
% \multicolumn{1}{c|}{Model} & FLOPs ↓ & Model Size ↓ \\ \midrule
% CP-360 (2018) & 105.8G & 360.0M \\ \hline
% TASED Net (2019) & 91.8G & 85.4M \\ \hline
% Martin (2020) & \underline{27.9M} & 11.7M \\ \hline
% ATSal (2021) & 277.2G & 383.6M \\ \hline
% SST\_Sal (2022) & \textbf{768.0K} & \textbf{225.4K} \\ \hline
% STSA Net (2023) & 193.7G & 643.9M \\ \hline
% Pred360 (2024) & 306.5M & \underline{1.0M} \\ \hline
% Ours & 1.21G & 14.13M \\ \bottomrule
% \end{tabular}
% \vspace{-1em}
% \end{table}
% }
Table~\ref{ablation-param} presents a comparison of model complexity and computational efficiency for recent 360-degree video saliency prediction methods, with results sourced from~\cite{Wan2024}~\cite{Yang2024}. Our proposed model is notably lightweight, comprising only 3.70 million parameters and requiring 14.13 MB of storage, while achieving a competitive computational cost of 1.21 GFLOPs. Based on Table~\ref{ablation-param}, and in comparison with other state-of-the-art approaches, our model has the lowest number of parameters and ranks fourth in terms of FLOPs and model size.
%In contrast, heavier architectures such as ATSal (383.6 MB) and CP-360 (360 MB), along with more recent approaches like 360Spred (7.59M parameters), are substantially larger, our model is over ten times smaller in size.

This compact design makes our method particularly well-suited for edge-side or client-side deployment, where computational resources and memory are limited (\textit{e.g.}, in VR headsets). Our approach strikes a balance between lightweight design, accuracy, and efficiency, enabling real-time saliency estimation directly on user devices.
Additionally, we measured the average inference time of our model to be $5.1\text{ms}$  per frame, corresponding to a throughput of approximately 196 frames per second (fps) on the hardware described in the experimental setup. This confirms the high efficiency of the model and provides strong evidence that it can operate seamlessly in real-time applications, like viewport prediction, without introducing noticeable latency.

\begin{table}[t]
\centering
\caption{Comparison of model complexity and computational efficiency across models. Best results are in \textbf{bold}, second-best are \underline{underlined}.}
\setlength{\tabcolsep}{.2em}
\renewcommand{\arraystretch}{1.1}
\begin{tabular}{l|c|c|c}
%\toprule
Model & Parameters (M) ↓ & FLOPs ↓ & Model Size ↓ \\
\hline %\midrule
Offline DHP~\cite{Xu2018_2} & 34.00 & -- & -- \\ 
%\hline
SalGAN360~\cite{Chao2018} & 31.78 & -- & -- \\ 
%\hline
Spherical U-Net~\cite{Zhang_ECCV_2018} & 12.30 & -- & -- \\ 
%\hline
SST-Sal~\cite{Edurne2022} & 8.32 & \textbf{768.0K} & \textbf{225.4K} \\ 
%\hline
SPVP360~\cite{Jie2022} & 7.72 & -- & -- \\ 
%\hline
360Spred~\cite{Yang2024} & \underline{7.59} & -- & -- \\ 
%\hline
CP360~\cite{cheng2018} & -- & 105.8G & 360.0M \\ 
%\hline
TASED Net~\cite{Min2019} & -- & 91.8G & 85.4M \\ 
%\hline
Martin~\cite{martin2020} & -- & \underline{27.9M} & 11.7M \\ 
%\hline
ATSal~\cite{Yasser2020} & -- & 277.2G & 383.6M \\ 
%\hline
STSA Net~\cite{Wang2021} & -- & 193.7G & 643.9M \\ 
%\hline
Pred360~\cite{Wan2024} & -- & 306.5M & \underline{1.0M} \\ 
\hline
\textbf{Ours} & \textbf{3.70} & 1.21G & 14.13M \\
%\bottomrule
\end{tabular}
\label{ablation-param}
\end{table}

\begin{table}[t]
\centering
\caption{Ablation study across datasets showing the impact of Center Bias through $\delta$ and $\beta$ components. Best results are in \textbf{bold}.}
\label{tab:ablation}
\renewcommand{\arraystretch}{1.2}
\setlength{\tabcolsep}{10pt}
\begin{tabular}{cc|c:c:c:c}
\multicolumn{1}{c}{$\delta(\cdot)$} & \multicolumn{1}{c|}{$\beta$} & \multicolumn{1}{c}{CC ↑} & \multicolumn{1}{c}{NSS ↑} & \multicolumn{1}{c}{KL ↓} & \multicolumn{1}{c}{AUC Judd ↑} \\
\hline
\multicolumn{6}{c}{\textbf{Sport360}} \\ 
\hline
\xmark & \xmark & 0.696 & 4.482 & 3.461 & 0.941 \\ 
\xmark & \cmark & 0.707 & 4.537 & 3.372 & 0.942 \\
\cmark & \xmark & 0.716 & 4.647 & 3.303 & 0.943 \\ 
\cmark & \cmark & \textbf{0.722} & \textbf{4.746} & \textbf{3.267} & \textbf{0.943} \\ 
% \midrule
\hline
\multicolumn{6}{c}{\textbf{PVS-HM}} \\ 
\hline
\xmark & \xmark & 0.768 & 3.649 & 1.254 & 0.933 \\ 
\xmark & \cmark & 0.790 & 3.827 & 1.139 & 0.935 \\ 
\cmark & \xmark & 0.776 & 3.720 & 1.120 & 0.936 \\ 
\cmark & \cmark & \textbf{0.807} & \textbf{3.946} & \textbf{1.042} & \textbf{0.937} \\ 
\hline
\multicolumn{6}{c}{\textbf{VR-EyeTracking}} \\ 
\hline
\xmark & \xmark & 0.570 & 2.780 & 2.461 & 0.910 \\ 
\xmark & \cmark & 0.581 & 2.884 & 2.464 & 0.911 \\ 
\cmark & \xmark & 0.586 & 2.992 & 2.432 & \textbf{0.912} \\
\cmark & \cmark & \textbf{0.593} & \textbf{3.016} & \textbf{2.334} & \textbf{0.912} \\
\end{tabular}
\label{ablation-biases}
\end{table}

\section{Ablation study}\label{sec:abltation_study}
%\subsection{spherical convolution}
%To evaluate the impact of using spherical convolution and pooling layers, we conducted tests on our model by replacing these layers in the decoder with traditional convolution and pooling layers, while keeping all other parameters unchanged. The results are presented in Table 2.

\subsection{Viewing Center Bias contribution}
Table~\ref{ablation-biases} presents an ablation study done across three datasets (Sport360, PVS-HM, and VR-EyeTracking) to evaluate the contribution of the \ac{CB} on our model through $\delta$ and $\beta$ components.

%On the Sport360 dataset, removing either \ac{IFCB} or \ac{CB} consistently leads to drops in performance across all metrics. The absence of both components results in the lowest performance, confirming their complementary roles. When both modules are included, the model achieves the best results. Additionally, the contribution of \ac{IFCB} has a greater impact on the model performance than that of \ac{CB} in this dataset.

%Similarly, for the PVS-HM dataset, the contributions of \ac{IFCB} and \ac{CB} are significant. When one of the modules is removed, the model performance degrades, and the best results are achieved when both components are incorporated. In this dataset, the contribution of \ac{CB} is greater than that of \ac{IFCB}.
The results clearly show that both $\delta$ and $\beta$ contribute to the performance improvement across all datasets, though with varying degrees of impact. Removing one of these components leads to performance degradation across all metrics, while removing both components results in the lowest overall accuracy. 

On the Sport360 dataset, the $\beta$ component has a smaller influence compared to $\delta$, indicating that this dataset exhibits a weaker long-range dependence on center bias. In contrast, the PVS-HM dataset shows a significant performance drop when the $\beta$ is removed, suggesting a stronger reliance on persistent center bias. These findings align with observations reported in~\cite{Chen_2023}, which highlight that PVS-HM demonstrates a more center-biased behavior across its videos than Sport360. On the other hand, Sport360 appears to rely more on $\delta$, as the center bias primarily occurs at the beginning of the videos, likely due to the characteristics of the content within this dataset.

In the VR-EyeTracking dataset, the benefits of both $\delta$ and $\beta$ are evident. Each component individually leads to improvements in evaluation metrics, especially in \ac{CC} and \ac{NSS}, while their combination consistently yields the strongest results, with a \ac{CC} of 0.593, \ac{NSS} of 3.016, and \ac{KL} of 2.334. Here, the contributions of $\delta$ and $\beta$ are nearly equal, indicating that the VR-EyeTracking dataset exhibits a balance between early frame-centered viewing behavior and sustained central attention patterns.

Overall, these results confirm that both $\delta$ and $\beta$ are important for modeling realistic viewing behavior. Specifically, $\delta$ aligns predictions with the natural tendency of users to focus on the center at the beginning of a video, while $\beta$ accounts for the global bias during viewing. Together, they provide complementary benefits, leading to consistently improved saliency predictions across all datasets. Furthermore, the contributions of $\delta$ and $\beta$ can vary from one dataset to another, highlighting the importance of integrating these components and incorporating them as learnable parameters within the model. This allows for adaptation based on the content of each dataset.

% \begin{table}[t]
% \centering
% \caption{Ablation study across datasets showing the impact of Center Bias through $\delta$ and $\beta$ components.}
% \label{tab:ablation}
% \renewcommand{\arraystretch}{1.2}
% \begin{tabular}{p{0.15\textwidth}:c:c:c:c}
% %\begin{tabular}{p{0.15\textwidth}cccc}

% \multicolumn{1}{c}{Component} & \multicolumn{1}{c}{CC ↑} & \multicolumn{1}{c}{NSS ↑} & \multicolumn{1}{c}{KL ↓} & \multicolumn{1}{c}{AUC Judd ↑} \\
% % \midrule
% \hline
% \multicolumn{5}{c}{\textbf{Sport360}} \\ 
% \hline
% w/o $\delta$ & 0.707 & 4.537 & 3.372 & 0.942 \\
% % \hline
% w/o $\beta$ & 0.716 & 4.647 & 3.303 & 0.943 \\ 
% % \hline
% w/o $\delta$, w/o $\beta$  & 0.696 & 4.482 & 3.461 & 0.941 \\ 
% % \hline
% Ours & \textbf{0.722} & \textbf{4.746} & \textbf{3.267} & \textbf{0.943} \\ 
% % \midrule
% \hline
% \multicolumn{5}{c}{\textbf{PVS-HM}} \\ 
% \hline
% w/o $\delta$ & 0.790 & 3.827 & 1.139 & 0.935 \\ 
% % \hline
% w/o $\beta$ & 0.776 & 3.720 & 1.120 & 0.936 \\ 
% % \hline
% w/o $\delta$, w/o $\beta$ & 0.768 & 3.649 & 1.254 & 0.933 \\ 
% % \hline
% Ours & \textbf{0.807} & \textbf{3.946} & \textbf{1.042} & \textbf{0.937} \\ 
% %\midrule
% \hline
% \multicolumn{5}{c}{\textbf{VR-EyeTracking}} \\ 
% % \midrule
% \hline
% w/o $\delta$ & 0.581 & 2.884 & 2.464 & 0.911 \\ 
% % \hline
% w/o $\beta$ & 0.586 & 2.992 & 2.432 & \textbf{0.912} \\
% % \hline
% w/o $\delta$, w/o $\beta$ & 0.570 & 2.780 & 2.461 & 0.910 \\ 
% % \hline
% Ours & \textbf{0.593} & \textbf{3.016} & \textbf{2.334} & \textbf{0.912} \\ %\bottomrule
% \end{tabular}
% \vspace{-1em}
% \label{ablation-biases}
% \end{table}

\begin{table}[t]
\centering
\caption{Learned values for $\alpha$ and $\beta$ parameters across datasets.}
\renewcommand{\arraystretch}{1.2}
\begin{tabular}{l|>{\centering\arraybackslash}p{2.5cm}|>{\centering\arraybackslash}p{2.5cm}}
% \toprule
\multicolumn{1}{c|}{\textbf{Dataset}} & $\alpha$ & $\beta$
 \\ 
\hline %\midrule
Sport360 & 908.5013 & 0.0030 \\ %\hline
PVS-HM & 573.5508 & 0.1773 \\ %\hline
VR-EyeTracking & 436.3028 & 0.1079 \\
% \bottomrule
\end{tabular}

\label{beta_alpha}
\end{table}

\begin{figure*}[t]
    \centering
    %\graphicspath{{images/}}
    \includegraphics[width=.9\linewidth]{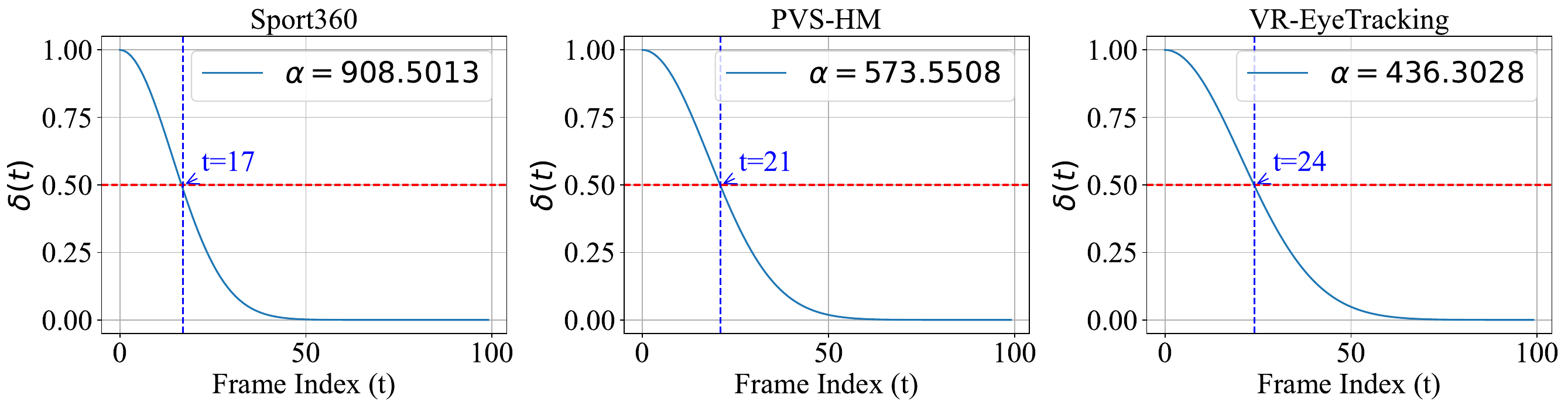}
    \caption{Temporal decay of the $\delta(t)$ function across frame indices for each dataset. The frame index at which $\delta(t)$ equals 0.5 is highlighted.} 
    %\label{Segmentation results obtained by feeding 360-degree equirectangular frames into the SegFormer model (MiT-B0)}
    \label{fig:weight_decay}
\end{figure*}

To further analyze the contribution of $\delta$ and $\beta$ components in our model, Table~\ref{beta_alpha} presents the final values of the parameters $\alpha$ and $\beta$ after convergence across the three datasets. As observed, the value of $\alpha$ decreases below its initial setting of 600 in the PVS-HM and VR-EyeTracking datasets, indicating that the relative contribution of $\delta$ increases in these datasets. In contrast, for the Sport360 dataset, $\alpha$ increases to 908.5, suggesting that the model reduces the influence of $\delta$. This behavior implies that PVS-HM and VR-EyeTracking exhibit stronger initial fixation tendencies, whereas Sport360 is less dependent on $\delta$.
To further illustrate the contribution of $\delta$, Figure~\ref{fig:weight_decay} plots the $\delta$ function across frame indices using the learned $\alpha$ values. The frame index corresponding to a $\delta$ of 0.5 has been highlighted. In Sport360, $\delta$ is equal to 0.5 at frame 17, compared to frame 21 in PVS-HM and frame 24 in VR-EyeTracking. These results confirm that VR-EyeTracking demonstrates the strongest $\delta$ dependence, followed by PVS-HM, while Sport360 shows the weakest.
Regarding the $\beta$ contribution, the parameter $\beta$ increases from its initial value of 0.15 to 0.17 in the PVS-HM dataset, indicating that the model strengthens the influence of the long-range center bias. In contrast, $\beta$ slightly decreases to 0.11 for VR-EyeTracking, showing a moderate but still relevant $\beta$ effect. For Sport360, however, $\beta$ drops to 0.003, suggesting that long-range center bias has minimal influence in this dataset compared to others.

\begin{table}[t]
\centering
\caption{Performance comparison across different loss configurations. Best results are in \textbf{bold}, second-best are \underline{underlined}.}
\setlength{\tabcolsep}{4pt}
\renewcommand{\arraystretch}{1.2}
\resizebox{\linewidth}{!}{%
\begin{tabular}{cccc|c:c:c:c}
% \toprule
$\mathcal{L}_{CC}$ & $\mathcal{L}_{KL}$ & $\mathcal{L}_{SMSE}$ & $\mathcal{L}_{BCE}$ & CC ↑ & NSS ↑ & KL ↓ & AUC Judd ↑ \\ 
\hline %\midrule
\cmark & \xmark & \xmark & \xmark & 0.780 & 3.786 & \textbf{0.967}  & 0.934 \\ 
% \hline
\xmark & \cmark & \xmark & \xmark & 0.778 & 3.725 & 1.236 & 0.935 \\ 
% \hline
\cmark & \cmark & \xmark & \xmark & 0.798 & 3.869 & 1.117 & \textbf{0.937} \\ 
% \hline
\cmark & \cmark & \cmark & \xmark & \underline{0.803} & \underline{3.927} & 1.064 & \underline{0.936} \\ 
\hdashline
\cmark & \cmark & \cmark & \cmark & \textbf{0.807} & \textbf{3.946} & \underline{1.042}  & \textbf{0.937} \\ 
% \bottomrule
\end{tabular}}
\label{ablation-losses}
\end{table}

%\subsection{Center bias contribution}
%\subsection{alpha cont.}
\begin{figure*}[ht!]
    \centering
    % -------- 
    \subfloat[\label{fig:a}]{\parbox{.3\textwidth}{%
          \includegraphics[width=\hsize]{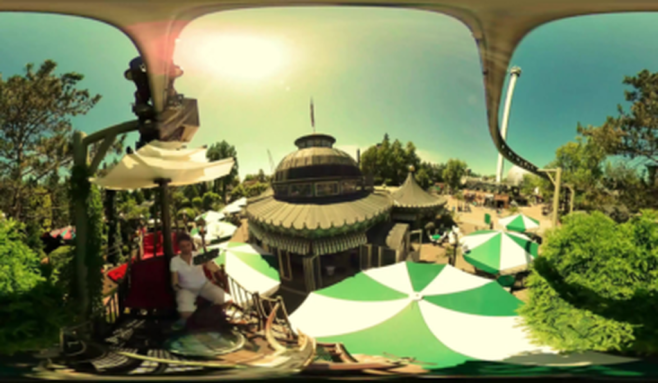}\vspace{2pt}
          \includegraphics[width=\hsize]{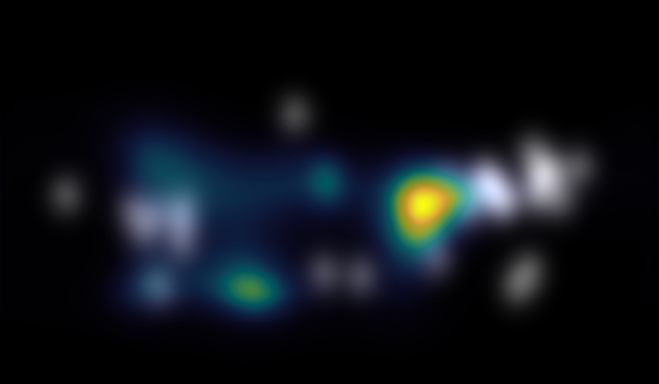}
         }}\hspace{0.2pt}
    \subfloat[\label{fig:b}]{\parbox{.3\textwidth}{%
          \includegraphics[width=\hsize]{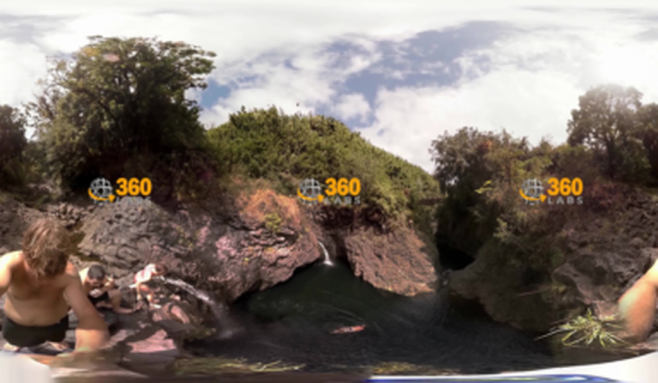}\vspace{2pt}
          \includegraphics[width=\hsize]{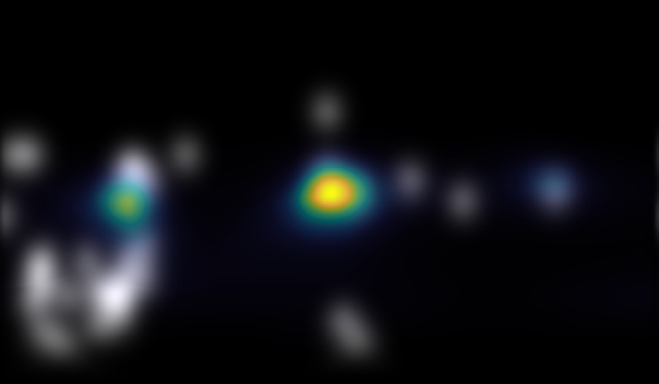}
         }}\hspace{0.2pt}
    \subfloat[\label{fig:c}]{\parbox{.3\textwidth}{%
          \includegraphics[width=\hsize]{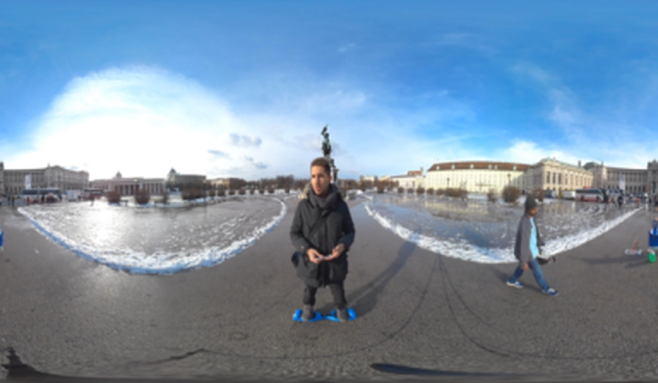}\vspace{2pt}
          \includegraphics[width=\hsize]{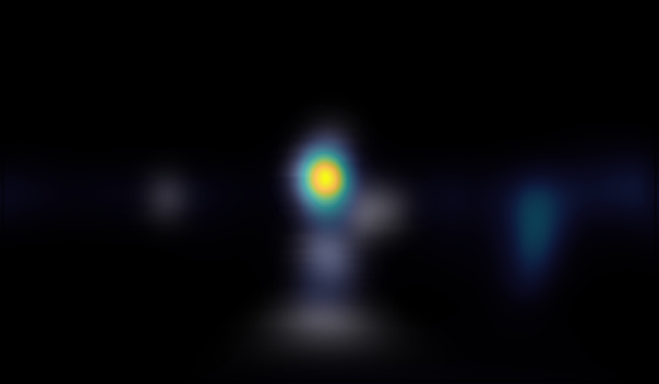}
         }}
        \hfill

    \vspace{4pt}

    \subfloat[\label{fig:d}]{\parbox{.3\textwidth}{%
          \includegraphics[width=\hsize]{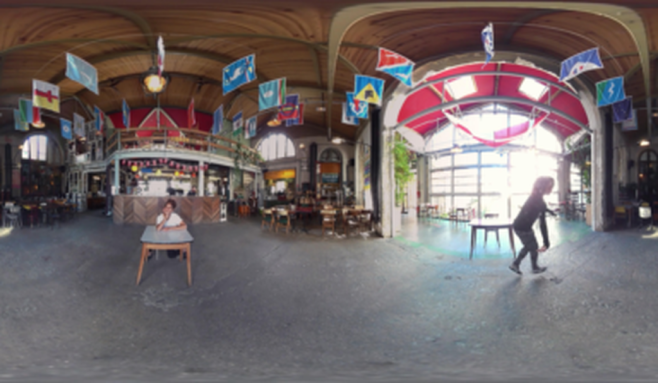}\vspace{2pt}
          \includegraphics[width=\hsize]{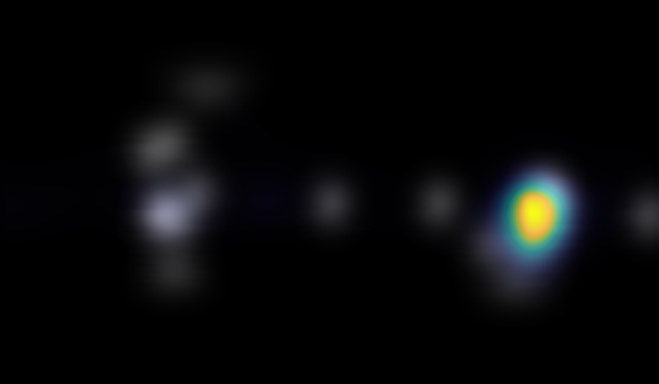}
         }}\hspace{0.2pt}
    \subfloat[\label{fig:e}]{\parbox{.3\textwidth}{%
          \includegraphics[width=\hsize]{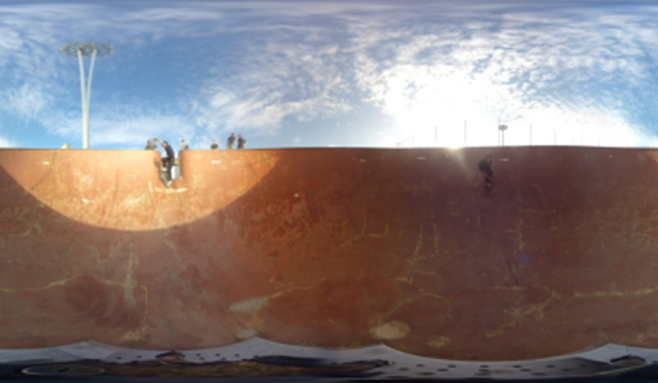}\vspace{2pt}
          \includegraphics[width=\hsize]{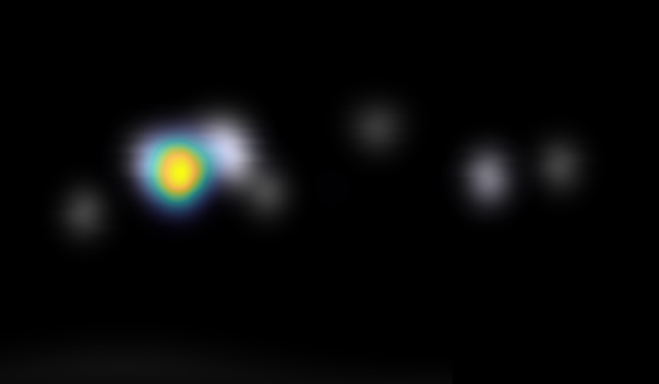}
         }}\hspace{0.2pt}
    \subfloat[\label{fig:f}]{\parbox{.3\textwidth}{%
          \includegraphics[width=\hsize]{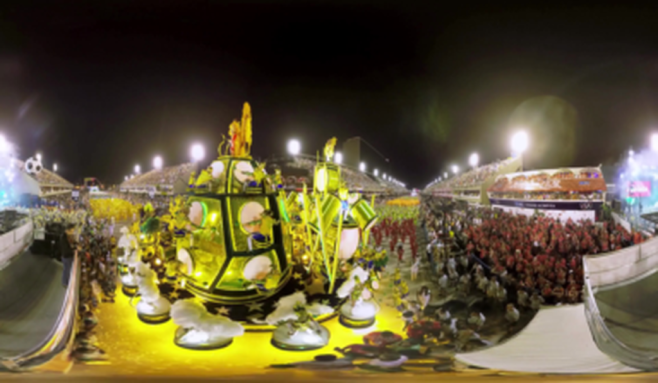}\vspace{2pt}
          \includegraphics[width=\hsize]{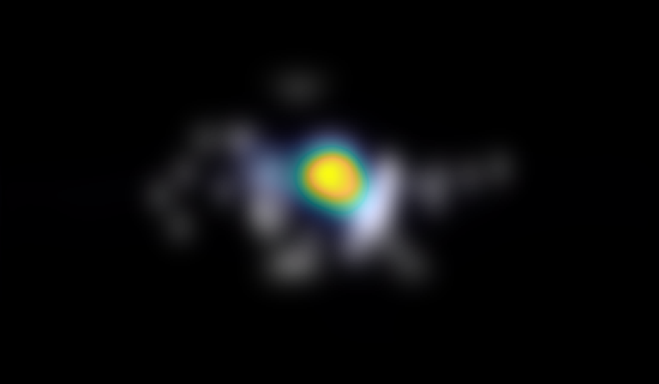}
         }}
        \hfill

    \caption{
        Qualitative analysis of challenging examples. Up: frames extracted from the three datasets; Down: ground-truth saliency maps (grayscale) with an overlay of the saliency maps predicted with the proposed method (parula colormaps).
    }
    \label{fig:Less-Accurate}
\end{figure*}

\subsection{Loss selection}
To highlight the contribution of each component in our loss function, we conducted an ablation study on the PVS-HM dataset by training and testing the model under different loss configurations (see Table~\ref{ablation-losses}). The results show that each loss plays an important role in improving the performance. Specifically, training the model with only the $\mathcal{L}_{CC}$ loss or the $\mathcal{L}_{KL}$ loss yields good results, but combining them ($\mathcal{L}_{CC} + \mathcal{L}_{KL}$) leads to a noticeable improvement across CC, NSS, and AUC metrics. Adding the $\mathcal{L}_{SMSE}$ further enhances the consistency of predictions, as reflected in higher \ac{CC} and \ac{NSS} scores and lower \ac{KL}. Finally, integrating the \ac{BCE} loss, $\mathcal{L}_{BCE}$, the model achieves the best overall performance, with the highest \ac{CC} (0.807), \ac{NSS} (3.946), and \ac{AUC-Judd} (0.937). Additionally, it results in a lower \ac{KL} of 1.042 compared to other loss functions, except for $\mathcal{L}_{CC}$, which has the lowest \ac{KL} among all the losses evaluated. A likely explanation is that optimizing solely with $\mathcal{L}_{CC}$ enforces a strong correlation between predicted and ground-truth maps, which closely align their distributions. This could explain the low value of the \ac{KL}, even though this loss does not yield the best overall performance across other metrics.

\subsection{Qualitative analysis of challenging examples}

Although our model outperforms existing state-of-the-art approaches on widely used datasets, several challenging cases with reduced prediction accuracy remain, as illustrated in Figure~\ref{fig:Less-Accurate}. The figure shows six frames extracted from the three datasets (upper rows) and the corresponding ground-truth saliency maps in grayscale with an overlay in parula colormap of the estimated saliency maps (lower rows).

In Figure~\ref{fig:a}, we show frame 546 from video 63 in the VR-EyeTracking dataset, recorded during an amusement park ride where the entire content changes rapidly from frame to frame with many points of interest inside the scene. Our model estimates salient areas primarily around the center, while the ground-truth maps reveal user fixations distributed across multiple regions, reflecting the difficulty of scenes with many simultaneous points of interest.

In Figure~\ref{fig:b}, we show frame 246 from video 24 of the VR-EyeTracking dataset. This video briefly displays an advertising logo (“360 Labs”), which disappears after a few seconds. Our model incorrectly detected the logo region as salient, overshadowing the main content. Moreover, the scene itself contained several competing points of interest, leading to user fixations scattered across the panorama, which our model struggled to predict accurately.

In Figure~\ref{fig:c}, depicting frame 53 of video 250 in the Sport360 dataset, two individuals appear in the scene. The ground-truth map shows strong focus on the central person, but our model distributed saliency between both individuals: one in the center moving his hands, and the other on the right who is moving across the scene. Comparable situations are also observed in the Sport360 dataset in Figure~\ref{fig:d} (frame 230 of video 301) and Figure~\ref{fig:e} (frame 286 of video 244), where the model prioritizes one salient region over another.

Finally, in Figure~\ref{fig:f}, we present frame 374 from the Rio Olympics video in the PVS-HM dataset. This video contains numerous people moving across the scene, creating complex motion patterns. While our model captured saliency regions to some extent, its predictions were less precise compared to the ground truth. Additionally, the video is composed of multiple stitched segments, causing frequent scene changes that further complicate saliency estimation.

\section{Future Directions}\label{sec:future_directions}
%To advance research in this area, we recommend creating new datasets with more diverse and carefully selected content. Specifically, datasets should avoid \textit{exploration videos} where no clear point of interest exists, or videos with an excessive number of points of interest scattered across the scene (e.g., VR-EyeTracking video 63, PVS-HM RioOlympics). Such cases often result in diffuse user fixations, making it difficult for deep learning models to learn meaningful patterns.

%Existing and future datasets also need systematic filtering:

%\begin{itemize}
%\item outlier users should be removed, as their irregular viewing behavior introduces noise.

%\item video content should remain consistent, avoiding concatenation of multiple unrelated scenes (e.g., PVS-HM RioOlympics, Sport360 video 218). This consistency is crucial because training saliency estimation models for video tasks depends on information from both current and previous frames, and abrupt content shifts can affect the learning process.

%\item logos or temporary overlays (e.g., the “360 Labs” logo in VR-EyeTracking video 24) should be excluded, since they mislead models into treating them as salient features.
%\end{itemize}

From a modeling perspective, there is still a strong need for more accurate, lightweight saliency estimation models suitable for real-time applications. This is especially critical for tasks such as viewport prediction, where high-quality saliency maps directly translate into better prediction performance as demonstrated in~\cite{Rondón2022}.
Furthermore, future models should integrate richer feature representations, such as spatial and temporal complexity cues or other types of information, to achieve faster convergence and enhance the accuracy of attention modeling across diverse 360-degree video content.

Another important direction is the improvement of the model's generalization capability and reliability. Indeed, the common practice in the evaluation of 360 degree video saliency estimation is to train and test on multiple independent datasets. Therefore, cross-domain performance remains unexplored. One possible way to address this limitation is to consider a wider range of dataset contents.
To achieve this, the creation of new large-scale datasets that capture a wider variety of scenes and are annotated with eye-tracking data from a larger and more diverse pool of users is needed. Such datasets would help mitigate bias and ensure that models remain robust under different viewing conditions and content types.

Finally, based on our observations of the contribution of center viewing bias, further studies and experiments are needed to better understand how this bias influences attention in 360-degree videos. Incorporating these insights into future models could lead to more accurate saliency estimation.

\section{Conclusion}\label{sec:conclusions}

In this paper, we propose SalFormer360, a transformer-based model for saliency estimation in 360-degree videos. By customizing the SegFormer backbone and designing a tailored decoder with additional priors, our model effectively shows high capability to predict the saliency maps for 360-degree videos. Experimental results across three large-scale benchmark datasets demonstrate that SalFormer360 consistently outperforms state-of-the-art methods, achieving improvements of up to 18.6\% in \ac{CC}. SalFormer360 is a lightweight model with low computational requirements, which makes it well-suited for real-time applications such as viewport prediction, and has the capability to be deployed on edge devices or client-side devices like head-mounted displays, enabling practical integration into immersive media systems.

%\vfill\pagebreak

%\clearpage
%\balance
%\bibliographystyle{IEEEbib}
%\bibliography{refs}
\bibliography{refs}

%\newpage
\vspace{-2cm}
\begin{IEEEbiography}[{\includegraphics[width=1in,height=1.25in,clip,keepaspectratio]{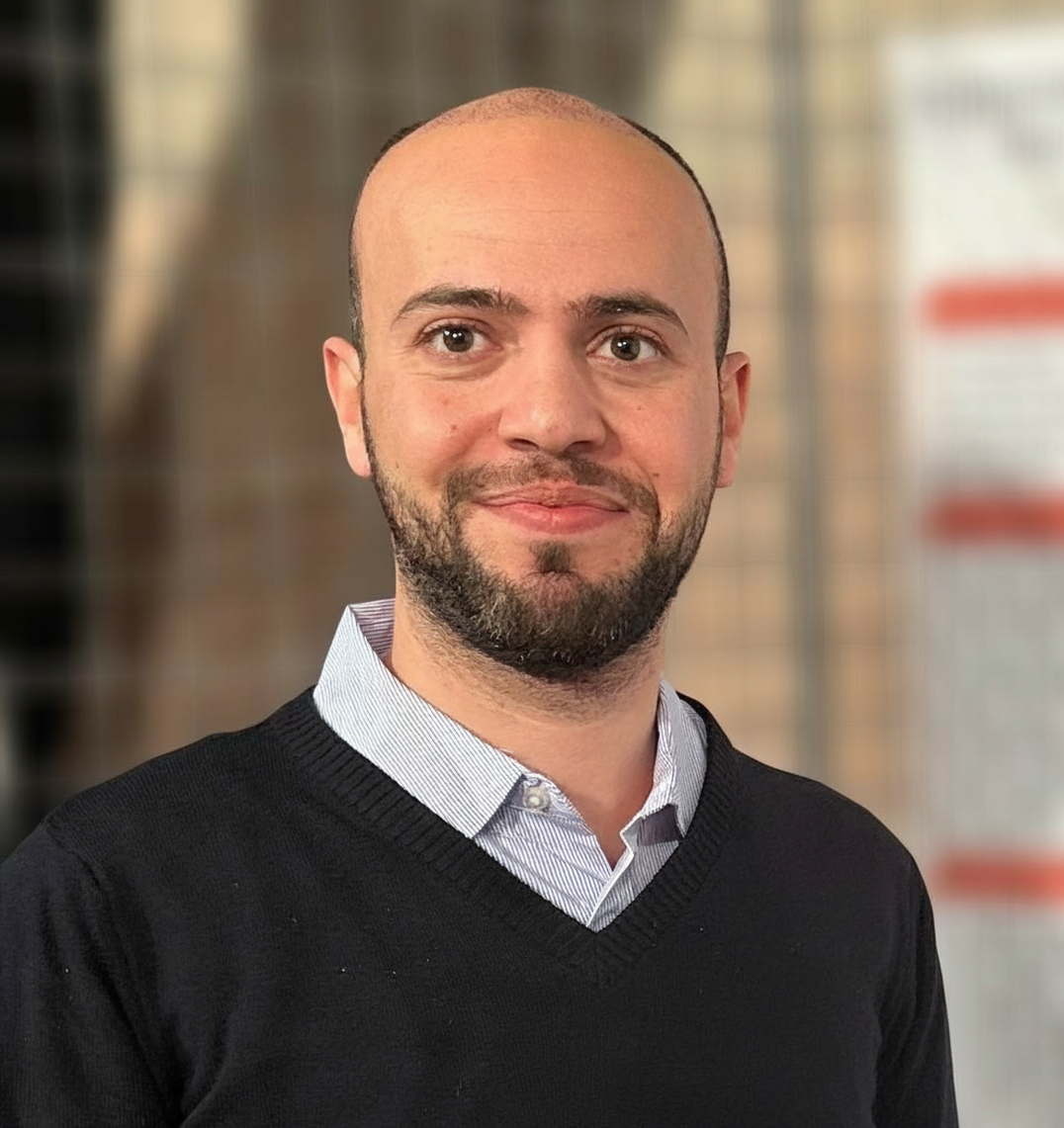}}]
{Mahmoud Z. A. Wahba} is a Ph.D. candidate at the University of Padova. He received his M.Sc. degree in ICT for Internet and Multimedia Engineering from the University of Padova in 2023 and his B.Sc. degree in Communications and Computer Engineering from Al-Azhar University in 2016. His research interests are in Computer Vision and Deep Learning, with a focus on immersive media streaming, especially for 360-degree video applications.
\end{IEEEbiography}
\vspace{-2cm}
\begin{IEEEbiography}[{\includegraphics[width=1in,height=1.25in,clip,keepaspectratio]{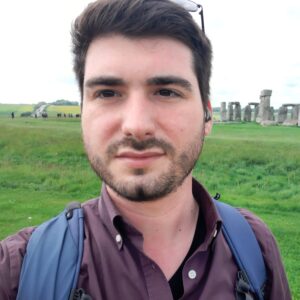}}]{Francesco barbato} is a Postdoctoral Researcher at the University of Padova, where he completed his Ph.D. in 2025. His doctoral research addresses limitations of machine learning models, focusing on dynamic, decentralized, and incremental environments, and training under data-limited conditions. His current work explores foundation model guidance and cross-architectural distillation to efficiently address these challenges on smaller models.
\end{IEEEbiography}
\vspace{-2cm}
\begin{IEEEbiography}[{\includegraphics[width=1in,height=1.25in,clip,keepaspectratio]{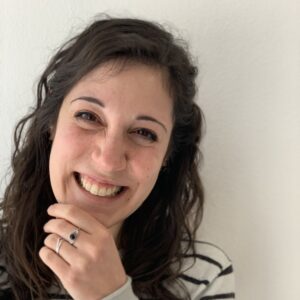}}]{Sara Baldoni} is Assistant Professor at the University of Padova from 2023. She received her PhD in Applied Electronics from Roma Tre University in 2022. Her main research interests are in multimedia signal processing, quality of experience, and immersive technologies. She is Area Editor for ELSEVIER Signal Processing: Image Communication. In addition, she is a member of the EURASIP Technical Area Committee on Visual Information Processing. 
\end{IEEEbiography}
\vspace{-2cm}
\begin{IEEEbiography}[{\includegraphics[width=1in,height=1.25in,clip,keepaspectratio]{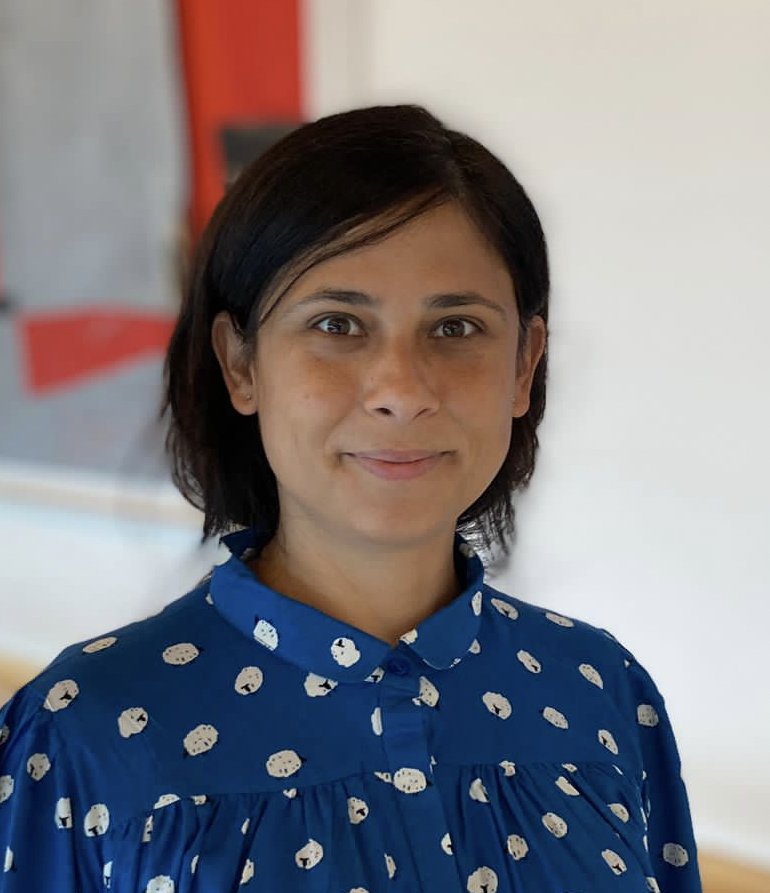}}]{Federica Battisti} (Senior Member, IEEE) is Associate Professor with the Department of Information Engineering, University of Padova. Her research interests include quality assessment for immersive media and multimedia signal processing. She is Editor in Chief for \textit{Signal Processing: Image Communication} (Elsevier), Chair of the IEEE SPS Italy Chapter, and Chair of the EURASIP Technical Area Committee on Visual Information Processing. 
\end{IEEEbiography}

% or


% \appendices

% % you can choose not to have a title for an appendix
% % if you want by leaving the argument blank
% \section{}

% % use section* for acknowledgment
% \section*{Acknowledgment}


\ifCLASSOPTIONcaptionsoff
  \newpage
\fi

\end{document}